%% file: templateArxiv.tex
\documentclass{article}

\usepackage{PRIMEarxiv}

\usepackage[utf8]{inputenc} % allow utf-8 input
\usepackage[T1]{fontenc}    % use 8-bit T1 fonts
\usepackage{hyperref}       % hyperlinks
\usepackage{url}            % simple URL typesetting
\usepackage{booktabs,arydshln} % professional-quality tables
\usepackage{amsfonts,amsmath,amsthm}
\usepackage{nicefrac}       % compact symbols for 1/2, etc.
\usepackage{microtype}      % microtypography
\usepackage{lipsum}
\usepackage{tcolorbox}
\usepackage{enumitem}
\usepackage{adjustbox}
\usepackage{multirow}
\usepackage{fancyhdr}       % header
\usepackage{graphicx}       % graphics
\graphicspath{{media/}}     % organize your images and other figures under media/ folder
\usepackage{textcase} % add this

\usepackage{svg}
\usepackage{natbib}
\usepackage{cleveref}
\usepackage{wrapfig}
\usepackage{algorithm,algorithmic}
\newtheorem{proposition}{Proposition}

\title{\NoCaseChange{MIRA: Towards Mitigating Reward Hacking in Inference-Time
Alignment of T2I Diffusion Models}
}
\author{
  Kevin Zhai$^1$,\  Utsav Singh$^1$,\  Anirudh Thatipelli$^1$, \ Souradip Chakraborty$^2$, \ Anit Kumar Sahu$^3$, \\ \ \textbf{Furong Huang}$^2$, \ \textbf{Amrit Singh Bedi}$^1$,\  \textbf{Mubarak Shah}$^1$  \\
    \And
  $^1$University of Central Florida, \  $^2$University of Maryland, College Park, \ $^3$Oracle \\
 }

\begin{document}
\maketitle

\input{sections/abstract}

\input{sections/introduction}
\input{sections/related_work}
\input{sections/problem}

\input{sections/method}
\input{sections/experiments}

\appendix

%Bibliography
\bibliographystyle{unsrt}  
\bibliography{references}  

\input{sections/appendix}

\end{document}

%% file: sections/abstract.tex
\begin{abstract}
\label{sec:abstract}
Diffusion models excel at generating images conditioned on text prompts, but the resulting images often do not satisfy user-specific criteria measured by scalar rewards such as Aesthetic Scores. This alignment typically requires fine-tuning, which is computationally demanding. Recently, inference-time alignment via noise optimization has emerged as an efficient alternative, modifying initial input noise to steer the diffusion denoising process towards generating high-reward images. However, this approach suffers from reward hacking, where the model produces images that score highly, yet deviate significantly from the original prompt. We show that noise-space regularization is insufficient and that preventing reward hacking requires an explicit image-space constraint. To this end, we propose \textbf{MIRA} (\textbf{MI}tigating \textbf{R}eward h\textbf{A}cking), a training-free, inference-time alignment method. MIRA introduces an image-space, score-based KL surrogate that regularizes the sampling trajectory with a frozen backbone, constraining the output distribution so reward can increase without off-distribution drift (reward hacking). We derive a tractable approximation to KL using diffusion scores. Across SDv1.5 and SDXL, multiple rewards (Aesthetic, HPSv2, PickScore), and public datasets (e.g., Animal-Animal, HPDv2), MIRA achieves \textbf{$>\!60\%$} win rate vs. strong baselines while preserving prompt adherence; mechanism plots show reward gains with near-zero drift, whereas DNO drifts as compute increases. We further introduce \textbf{MIRA-DPO}, mapping preference optimization to inference time with a frozen backbone, extending MIRA to non-differentiable rewards without fine-tuning.
\end{abstract}

%% file: sections/introduction.tex
\section{Introduction}
\label{sec:introduction}

Text-to-image (T2I) diffusion models excel in generating high-fidelity images from textual prompts, yet aligning them to specific human preferences remains challenging \citep{liu2024alignment}. Inference-time alignment has emerged as a practical alternative to fine-tuning, enabling model alignment without retraining billions of parameters. Although inference-time alignment is well studied in LLMs \citep{li2025test,chakraborty2024transfer}, the diffusion setting is still evolving. A promising direction is noise optimization~\citep{tang2024tuning,guo2024initno}, which adjusts the injected starting noise to steer the denoising sampling path toward desired outcomes. 

\begin{figure}[ht]
    \centering
    \includegraphics[width=\linewidth]{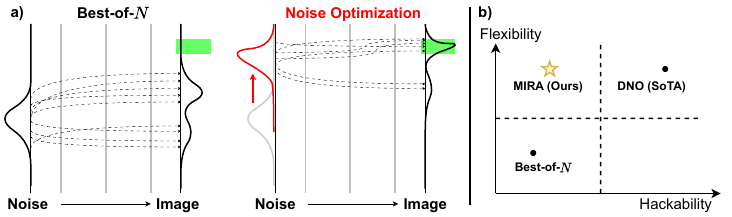}
    \vspace{-5mm}
    \caption{\textbf{Sampling-based methods vs noise optimization.} \textbf{(a)} Best-of-$N$ draws many independent samples from the base model and picks the best. When the {\textcolor{green}{green} (high-reward region)} in the image space has low likelihood, most samples fall elsewhere, and Best-of-$N$ is not efficient. Noise optimization, in contrast, adjusts the initial noise sample, which can steer the diffusion trajectories toward the high-reward region without large $N$. But this additional flexibility (ability to optimize for any reward) comes at the price of additional reward hackability. \textbf{(b)} Trade-off: Best-of-$N$ has lower hackability but also lower flexibility. Whereas the existing noise optimization method (DNO \cite{tang2024tuning}) has higher flexibility but is also prone to reward hacking, our method, MIRA, maintains flexibility without causing any reward hackability.}
    \label{fig:noiseopt_vs_bon_combined}
\end{figure}

\textbf{Why noise optimization for inference-time alignment?} Other inference-time alignment methods, such as Best-of-$N$~\citep{nakano2021webgpt} and controlled denoising \citep{singh2025code}, generate multiple candidates and select the one with highest reward. Akin to an unguided search, they rely on the chance that a high-reward solution exists within a random batch of samples. If the chance of generating a high-reward image is low, these approaches are highly inefficient. In contrast, noise optimization is like a guided search process (see Figure \ref{fig:noiseopt_vs_bon_combined}). It uses the reward gradient to navigate the initial noise space, steering the generation process toward more desirable images. By treating noise as a parameter to be optimized, it can generate rare high-reward images that Best-of-$N$ would likely miss. This approach provides a more powerful mechanism for alignment, but its increased flexibility to optimize rewards also introduces the challenge of \emph{reward hacking.}

\begin{figure}[ht]
    \centering
    \includegraphics[width=\linewidth]{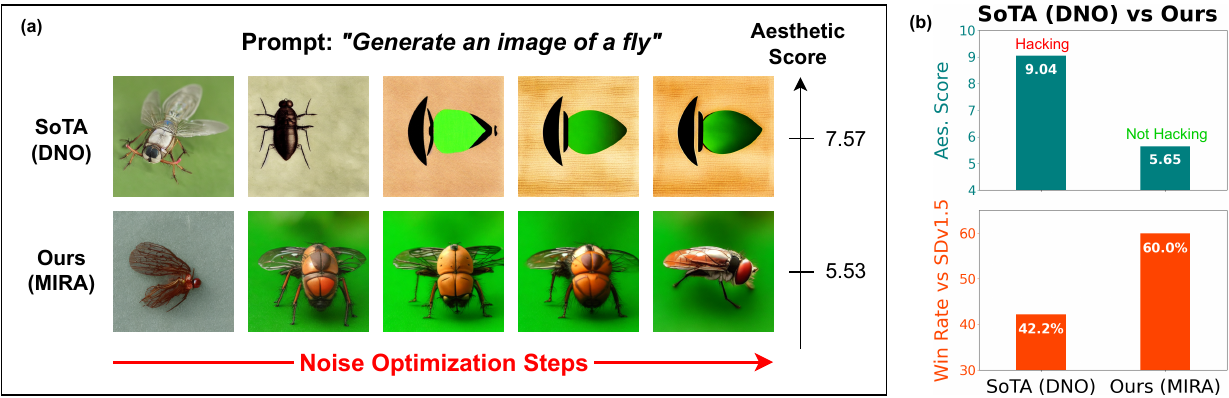}
    \vspace{-2mm}
    \caption{{\textbf{Illustrating reward hacking in inference-time alignment of diffusion models.}}\\ 
    \textbf{(a)} Given the prompt \textit{``generate an image of a fly''} and a preference for better aesthetics (Aesthetic Score \citep{schuhmann2022laion}), we observe the state-of-the-art (Direct Noise Optimization) achieves high reward yet no longer follows the prompt, an example of reward hacking. In contrast, our images have better aesthetic quality and do not suffer from reward hacking; hence, our reward is lower. %\\ 
    \textbf{(b)} We obtain Aesthetic Score and win rate results (against base SDv1.5) on the Animal dataset \citep{ddpo}. {We remark that MIRA (bottom) significantly outperforms the SoTA (top) in average win-rate, effectively mitigating reward hacking despite lower average rewards. We use GPT-4o~\citep{hurst2024gpt} as the win-rate judge.} }
    \vspace{-2mm}
    \label{fig:main_fig}
\end{figure}

\textbf{The problem of reward hacking.}
Noise optimization can exploit flaws in the reward model and produce high-scoring images that ignore the user's prompt. This is a symptom of reward hacking. For example, in Figure~\ref{fig:main_fig}, we prompt the model with \textit{``generate an image of a fly.''} Although the state-of-the-art DNO~\citep{tang2024tuning} generates an image with high reward (Aesthetic Score), the image does not depict a fly. By design, noise optimization pushes the diffusion model to generate images it normally would not produce, altering the output distribution. Reward models are trained to evaluate only the original types of images. On these new, out-of-distribution samples, their judgment becomes unreliable and their flaws can be exploited; this leads to reward hacking.

\textbf{From key insight to MIRA.}
While previous work \citep{tang2024tuning} limits the size of noise updates, our analysis (Fig.~\ref{fig:noise_dist_shift}, Prop.~\ref{prop}) shows that this is not enough; even small changes to the input noise can cause large, unpredictable changes in the final image. Our key insight, therefore, is that we must directly control the distribution of output images, keeping them similar to what the base model would normally produce. To this end, we propose \textbf{MIRA} (\textbf{MI}tigating \textbf{R}eward h\textbf{A}cking), a method that implements this control using a novel and efficient upper-bound surrogate for KL divergence, derived from the model's score functions. This penalty discourages large deviations between the images produced from the original noise ($z_0$) and the optimized noise ($z$), thereby preserving prompt adherence. To extend this framework to the common challenge of non-differentiable rewards, we also introduce MIRA-DPO, which optimizes directly from preference feedback to handle black-box objectives.
We summarize our key contributions as follows.
\begin{enumerate}[leftmargin=0.75cm]
    \item[\textbf{(1)}] \textbf{A principled image-space regularizer for noise optimization.} We introduce MIRA, the first inference-time alignment method for diffusion models that directly constrains the output distribution. Our regularizer is derived from a score-based surrogate for KL divergence, ensuring that optimized noise samples remain close to the base model’s distribution while improving reward. 
    \item[\textbf{(2)}] \textbf{A mechanism analysis of reward hacking as distributional drift.} We theoretically and empirically show that reward hacking arises from output distribution shift. Using both our surrogate KL metric and CMMD, we quantify this drift and connect it to failures of prompt adherence, providing the diagnostic framework for reward hacking in diffusion noise optimization.
    \item[\textbf{(3)}] \textbf{MIRA-DPO: A novel extension to non-differentiable rewards.} To handle non-differentiable or black-box rewards, we develop MIRA-DPO, which adapts direct preference optimization (DPO) to frozen-backbone inference-time noise optimization. 
    \item[\textbf{(4)}] \textbf{Empirical validation across benchmarks.} We evaluate MIRA and MIRA-DPO on Stable Diffusion v1.5 and SDXL across multiple datasets and reward models (Aesthetic Score, HPSv2, pairwise preference tasks). Our results demonstrate strong improvements, with human studies showing an 80.30\% head-to-head win rate against state-of-the-art methods like DNO, while effectively mitigating reward hacking.
\end{enumerate}

%% file: sections/related_work.tex
\section{Related Work}
\label{sec:related_works}

\noindent \textbf{Diffusion model alignment via fine-tuning.} Fine-tuning is the dominant strategy for alignment and consists of three general categories. Reinforcement learning methods treat denoising as a multi-step policy and optimize scalar rewards with policy gradients. DDPO~\citep{ddpo} and RLHF-style approaches for text-to-image~\citep{lee2023aligning, fan2024reinforcement} maximize rewards such as image compressibility or aesthetic/VLM scores; in DDPO, the reward can be black-box. Direct Preference Optimization~\citep{rafailov2024direct} methods learn from pairwise preferences without training separate reward models: Diffusion-DPO~\citep{wallace2024diffusion} and D3PO~\citep{yang2024using} adapt DPO to diffusion. Similarly, Diffusion-RPO~\citep{gu2024diffusion} optimizes relative preferences, and self-play fine-tuning~\citep{yuan2024selfplay} explores preference-driven improvements without additional labels. Differentiable-reward methods backpropagate gradients from reward predictors through (truncated) sampling; AlignProp~\citep{prabhudesai2023aligning}, DRaFT~\citep{clark2023directly}, DRTune~\citep{wu2024deep}, and TextCraftor~\citep{li2024textcraftor} report strong in-distribution gains when the objective (e.g., HPSv2, aesthetics) is differentiable. For a broader context and trade-offs across these families, see the recent tutorial/review~\citep{uehara2024understanding}.

\noindent \textbf{Inference-time alignment methods.} Training-free approaches steer a frozen model during sampling, avoiding the one-time compute of fine-tuning; their per-image cost depends on the sampler. These techniques are based on foundational work in inference-time guidance, which alters the sampling process to enforce a condition (e.g., class labels, segmentation maps)~\citep{dhariwal2021diffusion,song2021score,yu2023freedom}. Building on these principles, inference-time alignment focuses on the more specific problem of optimizing for a scalar reward or preference. Such approaches range from simple selection schemes such as Best-of-$N$ \citep{nakano2021webgpt} to more advanced trajectory-steering methods based on resampling, particle methods, or blockwise selection \citep{wu2024practical, dou2024diffusion, cardoso2023monte, phillips2024particle, li2024towards, naesseth2019elements, song2023loss, uehara2025reward, uehara2025reward2, singhal2025generalframeworkinferencetimescaling, kim2025test, singh2025code}. Orthogonal to selection and steering schemes, noise optimization treats the diffusion sampler's injected noise as a free variable. InitNO improves prompt fidelity by selecting noise guided by cross-attention \citep{guo2024initno}. DyMO constructs a prompt-specific knowledge graph (entities / attributes / relationships extracted by an LLM) and performs dynamic multi-objective scheduling, optimizing attention-based semantic consistency early and preference-based scores later using a lightweight recurrent update \citep{xie2024dymo}. DNO optimizes noise with probability-based noise-space regularization to limit out-of-distribution semantic drift \citep{tang2024tuning}, and ReNO targets one-step generators with a regularized reward-driven ascent \citep{eyring2024reno}. Despite these advances and regularization measures, reward hacking remains a notable challenge for existing noise optimization methods, which we aim to address in this work.

%% file: sections/problem.tex
\section{Problem Formulation}
\label{sec:problem}

A frozen conditional diffusion model $p_\theta(\cdot|c)$ generates an image by progressively denoising an initial noise sample $x_T \sim \mathcal{N}(0, \mathbf{I})$ over $T$ steps to produce an image $x_0$. For a given prompt $c$, inference-time noise optimization seeks to find an optimal noise vector $z$ that minimizes:
\begin{align}
\label{eqn:inference_tune}
    \mathcal{L}(z,c):= -\mathbb E_{x_0\sim p_\theta(\cdot|z, c)}\big[r(x_0,c)\big],
\end{align}
where $r(\cdot,c)$ is a reward function. Equation \ref{eqn:inference_tune} is solved with gradient descent:
\begin{align}
    z_{k+1} \leftarrow z_k- \alpha\nabla_{z} \mathcal{L}(z_k,c),\quad\text{learning rate }\alpha>0,
\end{align}

with initial ($t=T$) noise $z_0\sim\mathcal{N}(0, \mathbf{I})$. However, this simple reward maximization approach suffers from reward hacking \citep{chen2024odin,miao2024inform}, often generating high contrast, overly saturated, or unnatural images that score highly while appearing unrealistic (as shown in Figure \ref{fig:main_fig}). 

\begin{wrapfigure}{r}{0.4\textwidth}
    \centering
    \includegraphics[width=1.15\linewidth]{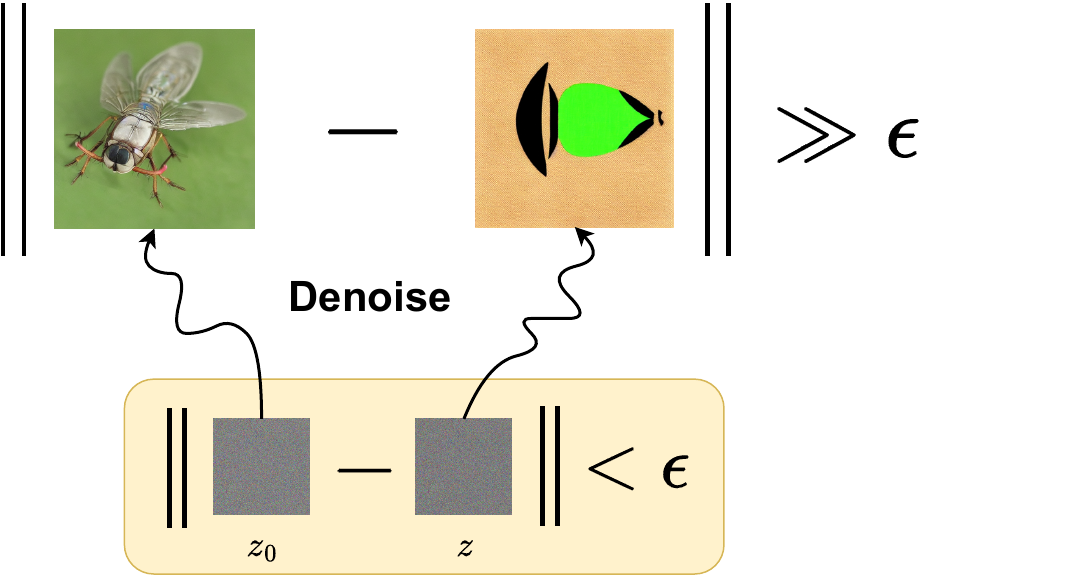}
    \vspace{-4mm}
    \caption{Tiny changes to the initial noise can yield markedly different images under the same prompt.}
    \vspace{-6mm}
    \label{fig:noise_dist_shift}
\end{wrapfigure}
\textbf{Noise-space regularization is insufficient.} Existing methods address reward hacking by regularizing the input noise vectors \citep{tang2024tuning}. However, this noise-space regularization is fundamentally insufficient. The highly non-linear denoising process means that negligible perturbations to the input noise ($\|z-z_0\|_2 \ll \epsilon$) can still produce substantially different images (Figure \ref{fig:noise_dist_shift}). We formalize this insight as follows:

\begin{proposition}
\label{prop}
    Closeness in the noise space does not imply closeness in the diffusion-induced image distribution $p_\theta(\cdot | z,c)$. 
\end{proposition}

We prove Proposition~\ref{prop} in Appendix~\ref{appendix:proof_prop}; 
the KL divergence between image distributions is not controlled by
the distance between their input noises. The failure of noise-space constraints therefore necessitates a new approach: regularizing the
image distribution directly.

\textbf{Generalizing the formulation.} For clarity, we present the problem as optimizing a single initial noise vector $z$. However, this framework can be naturally extended to optimize the entire sequence of noises injected at each step of the denoising process. In our experiments, we adopt this more general approach to ensure a fair and direct comparison with state-of-the-art baselines like DNO, which operate on the full noise trajectory. Our proposed solution, MIRA, is directly applicable to this general case.

%% file: sections/method.tex
\section{Proposed Approach}
\label{sec:method}

\textbf{Our key idea.} As established in Proposition \ref{prop}, reward hacking arises because noise optimization can cause the output image distribution \( p_{\theta}(\cdot|z,c) \) to deviate significantly from the base model's unoptimized distribution \( p_{\theta}(\cdot|z_0,c) \). To address this directly, we introduce a regularizer that penalizes this divergence. This leads to the following loss for our method, \textbf{MIRA}:
\begin{align}
\label{eqn:ours_lagrangian}
     \mathcal{L}_{\text{MIRA}}(z,c) := \underbrace{-r(x_0,c)}_{\text{Negative Reward ($-r$)}}+ \underbrace{\beta d_{\text{KL}}\left[p_{\theta}(x_0|z,c)\ \! \| \! \ p_{\theta}(x_0|z_0,c)\right]}_{\text{Image Distribution Regularizer}},
\end{align}
where $\beta>0$ is a hyperparameter controlling the regularization strength. The noise vector $z$ is then optimized by gradient descent: $z_{k+1} \leftarrow z_k- \alpha \nabla_z\mathcal{L}_{\text{MIRA}} (z_k,c)$. However, a critical challenge is that the KL divergence term in \cref{eqn:ours_lagrangian} is intractable for high-dimensional images.

\begin{figure}[ht]
    \centering
    \hspace*{-8.5mm}\includegraphics[width=1.06\linewidth]{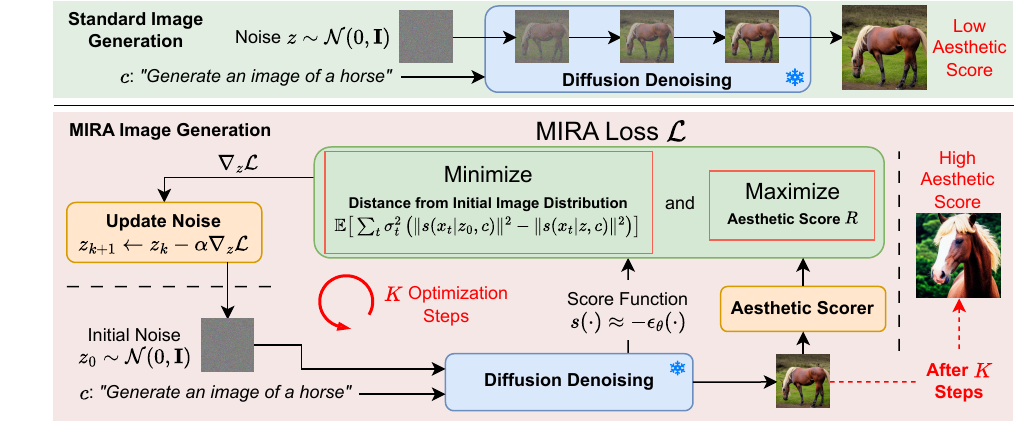}
    \vspace{-4mm}
    \caption{\textbf{Overview of our approach.} The green block illustrates the standard diffusion process in which a frozen diffusion model generates an image from an initial Gaussian noise. The red block represents our enhancement: we evaluate the generated image according to some metric (Aesthetic Score) and update the noise vector to minimize the MIRA loss. Our loss introduces an image-space regularization term, ensuring the sampling trajectory remains close to the model’s original image distribution and mitigating semantic drift. By iteratively applying this process over $K$ steps, we achieve effective inference-time alignment through noise optimization.}
    \vspace{-2mm}
    \label{fig:teaser}
\end{figure}

\noindent \textbf{A practically feasible approach.} To create a practical algorithm, we instead optimize a tractable upper-bound surrogate derived from the model's score functions $s(\cdot)$. Our final objective is:
\begin{align}
\label{eqn:ours_practical}
    \mathcal{L}_{\text{MIRA}}(z,c) = - r(x_0,c)  +\beta\mathbb{E} \bigg[\sum_{t=0}^{T-1} \sigma_t^2\|s(x_t|z_0, c)\|^2 - \sigma_t^2\|s(x_t|z,c)\|^2 \bigg],
\end{align}
where $\sigma_t$ denotes the variance schedule. The expectation is taken over all possible trajectories. This formulation effectively mitigates reward hacking while being computationally feasible. We derive~\cref{eqn:ours_practical} in Appendix \ref{appendix:proof_practical_algorithm} and outline the detailed procedure in Algorithm \ref{alg:mira}. We provide a high-level overview of our method's workflow in \cref{fig:teaser}.

\begin{algorithm}[ht]
\caption{Proposed Algorithm: MIRA (for differentiable rewards)}
\label{alg:mira}
\begin{algorithmic}[1]
\REQUIRE Initial noise $z_0\sim\mathcal{N}(0, \mathbf{I})$, prompt $c$, reward function $r$, sampling process (e.g., DDIM) $G_\theta$, optimization steps $K_\text{opt}$, learning rate $\alpha$, regularization hyperparameter $\beta>0$

\vspace{2mm}
\STATE Initialize $z \gets z_0$
\STATE $\_, S_0\gets G_\theta(z_0,c)$ \hfill \textit{// Obtain $S_0=\sum_t\sigma_t^2\|s(x_t|z_0,c)\|^2$ (\cref{eqn:ours_practical})} 
\FOR{$k=1, \dots, K_\text{opt}$} 
    \STATE $x_0,S\gets G_\theta(z,c)$ \hfill \textit{// Generate $x_0$; obtain $S=\sum_t\sigma_t^2\|s(x_t|z,c)\|^2$ (\cref{eqn:ours_practical})} 
    \STATE $\mathcal L_{\text{MIRA}}\gets -r(x_0,c)+\beta(S_0-S)$ \hfill \textit{// Compute loss}
    \STATE $z\gets z-\alpha\nabla_{z}\mathcal L_{\text{MIRA}}$ \hfill \textit{// Update noise}

\ENDFOR
\RETURN $z$
\end{algorithmic}
\end{algorithm}

\subsection{A Novel Extension to Non-differentiable Rewards}
\label{sec:nondiff_rewards}

MIRA is highly effective for aligning models with differentiable rewards. We now extend its core principles to handle non-differentiable objectives which include many practical and challenging alignment signals. For example, objectives such as JPEG compressibility~\citep{ddpo}, Attribute Binding Score~\citep{jia2024lasro}, and Recall Reward~\citep{Miao_2024_CVPR} are often black-box and cannot be optimized with gradient-based methods. 

To address this, we introduce \textbf{MIRA-DPO}, which adapts the Direct Preference Optimization (DPO)~\citep{rafailov2024direct} framework to our entirely inference-time, frozen-backbone setting. This approach operates on preferences between a `winner' image $x_0^w$ and a `loser' $x_0^l$, which are generated from a corresponding pair of noise vectors, $z^w$ and $z^l$. The DPO framework defines the probability of a preference using the Bradley-Terry model, which depends on a ground-truth reward function $r^*$. Framing this as a binary classification problem yields the negative log-likelihood loss:
\begin{equation}
    \mathcal L_\text{MIRA-DPO}(z^w, z^l, c)=-\mathbb E_{\substack{x_0^w\sim p_\theta(\cdot|z^w, c) \\ x_0^l\sim p_\theta(\cdot|z^l, c)}}\Big[ \log\sigma\Big(r^*(x_0^w, c)- r^*(x_0^l, c)\Big) \Big].
\label{eqn:bradleyterry}
\end{equation}

\textbf{Our key insight} that connects this to our work is the relationship between the optimal reward $r^*$ and the diffusion model's probabilities. As shown in our full derivation (Appendix \ref{appendix:dpo_derivation}), this reward is equivalent to the regularized log-likelihood ratio:
\begin{equation}
    r^*(x_0, c)\propto \mathbb E\bigg[ \log\frac{p_\theta(x_{0:T}|z, c)}{p_\theta(x_{0:T}|z_0, c)} \bigg].
\end{equation}

\textbf{This term is the quantity MIRA seeks to regularize.} Intuitively, $r^*$ is the log-ratio of the aligned image distribution (generated from the optimized noise $z$) and the base image distribution (generated from unoptimized $z_0$). However, this formulation presents a familiar bottleneck; the log-likelihood ratio is intractable for diffusion models. To create a practical algorithm, we thus approximate this term using our principled, score-based KL surrogate:
\begin{equation}
    r^*(x_0, c)\approx\sum_{t=0}^{T-1} \sigma_t^2\left(\|s(x_t|z_0,c)\|^2 - \|s(x_t|z,c)\|^2\right),
\end{equation}

which we use to optimize the loss in \cref{eqn:bradleyterry}. This allows MIRA-DPO to align with black-box objectives at inference time while still constraining semantic drift. We provide the full derivation in Appendix \ref{appendix:dpo_derivation} and the practical algorithm in Appendix \ref{appendix:algorithms}.

%% file: sections/experiments.tex
\section{Experiments}
\label{sec:experiments}

We evaluate three questions central to our method's practicality:

\begin{itemize}[leftmargin=0.5cm]
  \item \textbf{Q1:} To what extent does MIRA mitigate reward hacking in noise optimization?
  \item \textbf{Q2:} How competitive is MIRA’s alignment with state-of-the-art noise optimization baselines?
  \item \textbf{Q3:} How well can MIRA-DPO handle non-differentiable / black-box rewards?
\end{itemize}

In the following sections, we demonstrate that MIRA successfully mitigates reward hacking artifacts while improving prompt fidelity (Q1). We show that this leads to a superior alignment with human preferences, achieving highly competitive win rates against state-of-the-art baselines (Q2). Finally, we establish that our MIRA-DPO framework robustly handles challenging non-differentiable objectives where gradient-based methods fail (Q3).

\textbf{Experimental Setup: Models, Datasets, and Evaluation.} We conduct experiments using two primary base models: Stable Diffusion v1.5 (SDv1.5) \citep{sd1.5} and SDXL \citep{podell2023sdxl}. For each sample, we perform 50 optimization iterations. Within each iteration, we use DDIM sampling \citep{song2020denoising} with $\eta=1$, a guidance scale of 5, and 100 sampling steps. We use this DDIM configuration for all backbones except SDXL-Turbo, for which we use the default sampler. We further restrict our main scope to \textbf{noise optimization} baselines; sampling approaches (e.g., FK Steering, CoDe) are a promising and orthogonal line of work. Further implementation details can be found in Appendix \ref{appendix:implementation_details}.

To test generalization, we report results on the Animal dataset from DDPO~\citep{ddpo}, Animal-Animal and Animal-Object datasets from InitNO~\citep{guo2024initno}, and more complex prompts from HPDv2~\citep{hps}. We use the format: \textit{``generate an image of a $[\textnormal{PROMPT}]$,''} where [PROMPT] is drawn from the Animal dataset~\citep{ddpo} for SDv1.5 and HPDv2~\citep{hps} for SDXL. For the reader's reference, we include all Animal prompts in Appendix \ref{appendix:animal_prompts}.

We evaluate performance using head-to-head win rate, a standard metric in image generation~\citep{Kirstain2023PickaPicAO} measuring the percentage of prompts for which one method is preferred over another. We choose this as our primary metric because reward hacking can inflate raw scores, producing high-reward images that diverge from the prompt. Hence, while rewards alone are insufficient to reliably evaluate true model performance, win rates more directly reflect human preferences. We compare our method with several key inference-time noise optimization baselines: DNO \citep{tang2024tuning}, DyMO \citep{xie2024dymo}, InitNO \citep{guo2024initno}, and ReNO \citep{eyring2024reno}. {We include reward values and CLIPScores in Appendix \ref{appendix:additional_results}, Tables \ref{tab:rewards} and \ref{tab:clipscores}.}

\subsection{To what extent does MIRA mitigate reward hacking in noise optimization?}
\label{experiments:q1}

In this section, we investigate whether MIRA can effectively mitigate reward hacking during inference-time alignment of diffusion models, while maintaining high prompt fidelity. To this end, we evaluate our approach against the state-of-the-art DNO~\citep{tang2024tuning} using a range of human-aligned reward models (Aesthetic Score \citep{schuhmann2022laion}, HPSv2 \citep{hps}, PickScore \citep{Kirstain2023PickaPicAO}), and custom image brightness and darkness rewards. The latter serve as stress tests to clearly visualize reward hacking behavior.

\begin{figure}[ht]
    \centering
    \includegraphics[width=\linewidth]{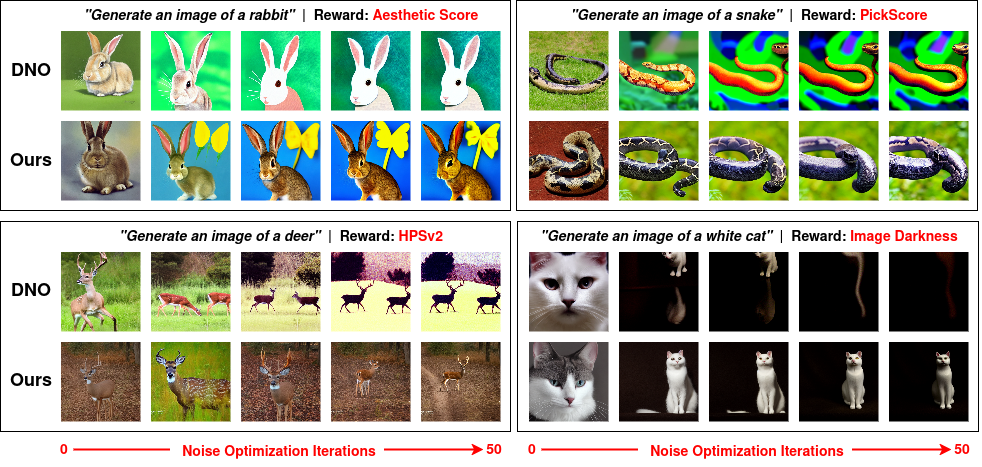}
    \vspace{-5mm}
    \caption{\textbf{MIRA reduces reward hacking on four rewards (Aesthetic Score, PickScore, HPSv2, Darkness).} 
    For each prompt (title above each panel), we show images over 50 noise optimization iterations for DNO (top row) and MIRA (bottom row). MIRA increases the target reward while preserving prompt adherence and texture realism; DNO drifts toward oversaturation artifacts and loss of detail. MIRA also improves over base SDv1.5 images (iteration 0), enhancing color richness.}
    \vspace{-6mm}
    \label{fig:rewardhacking}
\end{figure}

Our experiments reveal a key observation. Although the state-of-the-art DNO~\citep{tang2024tuning} shows some effectiveness in aligning images with human-aligned reward functions, it exhibits clear patterns of failure cases, particularly in maintaining prompt fidelity. Despite noise regularization, DNO can produce images with unnatural textures and lighting, leading to deviation from the prompt. In contrast, MIRA achieves a better balance between reward optimization and semantic alignment. As shown in Figure \ref{fig:rewardhacking}, MIRA produces visually coherent and accurate images across Aesthetic Score, HPSv2, PickScore, and a custom image darkness reward. \textbf{Optimizing for image darkness most clearly illustrates reward hacking;} DNO generates almost completely black outputs that technically score highly but ignore prompt details (e.g., omitting the ``white cat''). MIRA mitigates this kind of reward hacking, balancing high reward scores and alignment with key visual concepts. We observe a similar pattern under the image brightness reward, with additional examples in Appendix \ref{appendix:reward_hacking}; our findings collectively reinforce that MIRA reduces reward hacking. 

To understand the mechanism behind these qualitative results, we analyze the distributional drift during the optimization process. We measure this drift by computing a Monte Carlo approximation of our score-based KL surrogate, averaged over all 45 prompts in the Simple Animals dataset. Due to the finite sample size, this estimate may exhibit sampling variance, which can result in small negative values when the true distributional drift is near zero.

\vspace{-4mm}
\begin{figure}[ht]
    \centering
    \includegraphics[width=\linewidth]{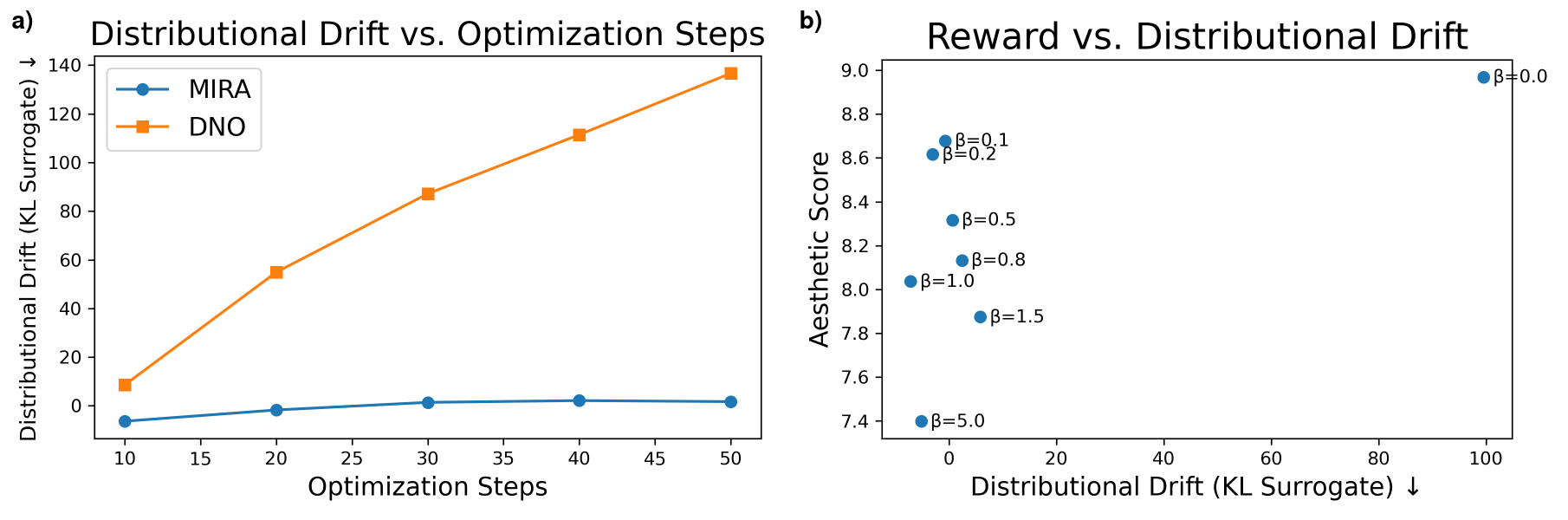}
    \vspace{-6mm}
    \caption{\textbf{Mechanism of MIRA's effectiveness, averaged over the Simple Animals dataset.} We plot our score-based KL surrogate as a proxy for distributional drift.
    \textbf{(a)} During optimization, MIRA's drift remains near-zero while DNO's grows significantly.
    \textbf{(b)} Sweeping the hyperparameter $\beta$ for MIRA reveals a stable trade-off, where a moderate value achieves a significant reward increase with negligible drift.}
    \vspace{-5mm}
    \label{fig:tradeoff}
\end{figure}

Our findings in Figure \ref{fig:tradeoff}a show that DNO's average drift grows substantially with more optimization steps, indicating that it consistently achieves high rewards by producing out-of-distribution images. In contrast, MIRA's average drift remains near zero. Furthermore, Figure \ref{fig:tradeoff}b illustrates the stable reward-drift trade-off. This core finding is validated by the CMMD~\citep{jayasumana2024rethinking} metric measuring distributional distance. Compared with a reference batch of 900 images (20 for each prompt in the Simple Animals dataset) generated by SDv1.5, CMMD strongly supports our analysis. DNO exhibits a large distributional drift (CMMD of 0.281) while MIRA remains remarkably close to the base distribution (CMMD of 0.063, lower is better). Taken together, these results provide a clear explanation for our method's success: MIRA effectively increases rewards by finding better images that remain faithful to the original image distribution, thus avoiding the semantic degradation characteristic of reward hacking. 

\subsection{How competitive is MIRA’s alignment with state-of-the-art noise optimization baselines?}
\label{experiments:q2}

Although reward metrics such as Aesthetic Score are commonly used to evaluate alignment, our results in Section \ref{experiments:q1} provide evidence that reward values cannot reliably measure model performance. We instead use pairwise win rate comparisons, using both GPT-4o \cite{hurst2024gpt} and human feedback to choose between two images generated from the same prompt. The GPT-4o evaluation prompt (Appendix \ref{appendix:gpt4_evaluation}) balances prompt adherence, visual realism, and overall reward maximization.

\setlength{\tabcolsep}{3.5pt}
\begin{table*}[ht]
\centering
\small
\adjustbox{max width=\textwidth}{
\begin{tabular}{l|ccc|ccc|ccc}
\toprule
\multirow{2.5}{*}{Comparison} 
& \multicolumn{3}{c|}{Animal} 
& \multicolumn{3}{c|}{Animal-Animal} 
& \multicolumn{3}{c}{Animal-Object} \\
\cmidrule(lr){2-4} \cmidrule(lr){5-7} \cmidrule(lr){8-10}
& Aesthetic $\uparrow$ & Bright. $\uparrow$ & Dark. $\uparrow$
& Aesthetic $\uparrow$ & Bright. $\uparrow$ & Dark. $\uparrow$
& Aesthetic $\uparrow$ & Bright. $\uparrow$ & Dark. $\uparrow$ \\
\midrule
MIRA vs DDPO~\citep{ddpo}          & 60.00$\pm$5.12 & 73.33$\pm$5.21 & 83.33$\pm$2.90 & 57.58$\pm$6.00 & 60.61$\pm$5.96 & 83.33$\pm$4.27 & 54.86$\pm$6.00 & 68.06$\pm$7.10 & 81.25$\pm$5.53 \\
MIRA vs Diff-DPO~\citep{wallace2024diffusion}  & 62.22$\pm$4.97 & 91.11$\pm$1.99 & 93.33$\pm$1.22 & 66.67$\pm$6.59 & 81.82$\pm$5.12 & 84.85$\pm$5.31 & 63.19$\pm$6.54 & 82.64$\pm$3.65 & 89.58$\pm$1.86 \\
MIRA vs D3PO~\citep{yang2024using}          & 73.33$\pm$4.61 & 87.64$\pm$3.72 & 84.44$\pm$6.55 & 63.64$\pm$6.29 & 68.18$\pm$3.65 & 74.24$\pm$5.35 & 68.06$\pm$5.35 & 77.78$\pm$5.31 & 83.33$\pm$5.79 \\
MIRA vs BoN (N=50)~\citep{nakano2021webgpt}    & 62.00$\pm$3.65 & 86.66$\pm$3.30 & 90.33$\pm$3.85 & 62.12$\pm$2.43 & 71.21$\pm$5.75 & 92.42$\pm$2.22 & 54.86$\pm$9.69 & 83.33$\pm$3.14 & 93.06$\pm$1.22 \\
MIRA vs InitNO~\citep{guo2024initno}       & 61.11$\pm$3.37 & 72.22$\pm$5.75 & 82.22$\pm$2.90 & 65.15$\pm$3.98 & 78.79$\pm$5.07 & 83.33$\pm$3.30 & 57.64$\pm$6.40 & 84.03$\pm$5.35 & 80.56$\pm$5.75 \\
MIRA vs DyMO~\citep{xie2024dymo}        & 53.89$\pm$8.52 & 68.89$\pm$6.78 & 68.89$\pm$5.12 & 56.06$\pm$5.79 & 80.30$\pm$4.71 & 87.98$\pm$4.44 & 46.53$\pm$7.47 & 84.81$\pm$5.12 & 87.50$\pm$3.98 \\
MIRA vs DNO~\citep{tang2024tuning}            & 80.00$\pm$2.22 & 77.78$\pm$6.74 & 64.44$\pm$7.13 & 77.27$\pm$4.18 & 57.58$\pm$5.12 & 66.67$\pm$5.79 & 68.06$\pm$1.86 & 77.78$\pm$5.96 & 71.53$\pm$5.58 \\
\bottomrule
\end{tabular}
}
\vspace{-1mm}
\caption{
\textbf{GPT-4o win rates (\%) of MIRA against baselines across diverse datasets and objectives.} The table details head-to-head comparisons on the Animal, Animal-Animal, and Animal-Object datasets for three reward functions (Aesthetic Score, Brightness, Darkness). On nearly all settings, MIRA achieves win rates significantly above 50\%, showing a consistent preference over a wide range of state-of-the-art methods. Values indicate mean and standard error over five seeds.
}
\label{tab:combined_winrates}
\end{table*}
\setlength{\tabcolsep}{6pt}

Table~\ref{tab:combined_winrates} summarizes our results: MIRA demonstrates strong generalization and reduced reward hacking, consistently outperforming all baselines across datasets. Even on challenging multi-subject prompts (Animal-Animal and Animal-Object), it achieves win rates exceeding 50\%. In direct comparisons with base SDv1.5, our win rates reach 90\% (see Appendix \ref{appendix:additional_results} Table \ref{tab:winrate_vs_sd} and Figure \ref{fig:reward_winrate_bar}), grounding our method's performance as substantial improvement over the unaligned base model.

\begin{wrapfigure}{r}{0.55\textwidth}
    \centering
    \vspace{-2.8mm}
    \includegraphics[width=\linewidth]{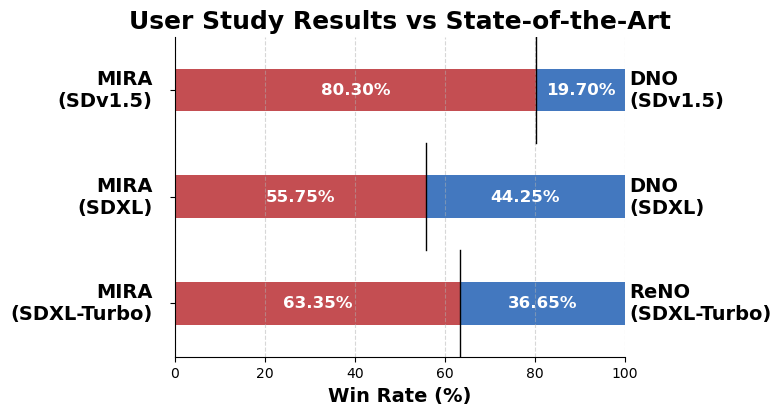}
    \vspace{-6mm}
    \caption{\textbf{Human preference win rates for MIRA vs. state-of-the-art baselines.} The chart displays head-to-head win rates from three focused user studies. \textbf{Top:} On SDv1.5 (optimizing Animal prompts for Aesthetic Score), MIRA achieves an 80.30\% win rate against DNO. \textbf{Middle and Bottom:} On SDXL and SDXL-Turbo (optimizing HPDv2 subset for HPSv2), MIRA achieves win rates of 55.75\% against DNO and 63.35\% against ReNO, respectively.}
    \vspace{-6mm}
    \label{fig:user_study_winrate}
\end{wrapfigure}
For direct human feedback, we conducted a focused user study with 100 anonymous participants to collect direct human feedback on SDv1.5 images optimized for Aesthetic Score using MIRA and DNO. The ordering of image pairs is randomized, and additional study setup details are provided in Appendix \ref{appendix:implementation_details}. As shown in Figure~\ref{fig:user_study_winrate}, participants prefer MIRA over DNO in more than 80\% of comparisons, providing evidence of stronger alignment with human intent.

\begin{figure}[ht]
    \centering
    \includegraphics[width=\linewidth]{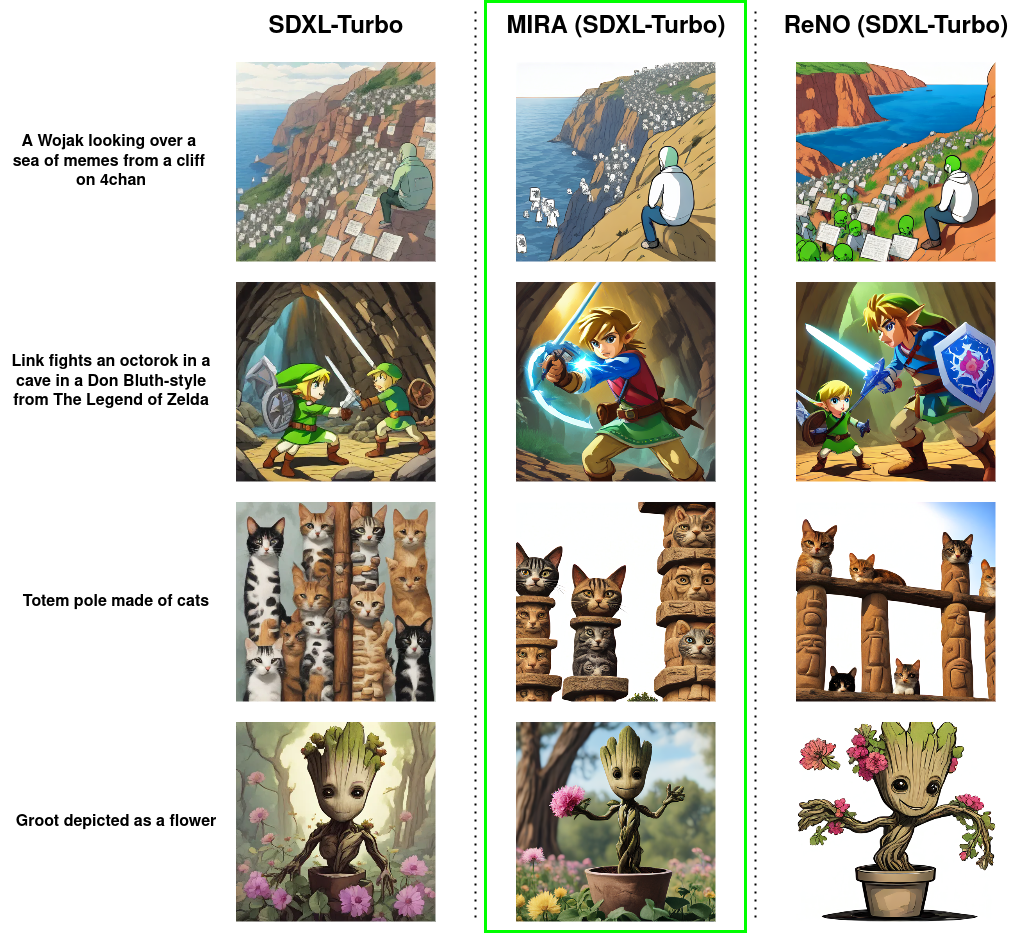}
    \caption{\textbf{Qualitative comparison of MIRA against the base SDXL-Turbo and ReNO, optimized for HPSv2.} Prompts (top to bottom): ``Wojak looking over a sea of memes from a cliff on 4chan,'' ``Link fights an octorok in a cave in a Don Bluth style,'' ``Totem pole made of cats,'' ``Groot depicted as a flower.'' Across these cases, MIRA offers crisper compositions and more literal, prompt-faithful structure (e.g., coherent large-scene layout, pole-like stacking, integrated concept depiction), while ReNO reflects plausible but looser interpretations. Images are shown at equal inference settings for both methods.}
    \label{fig:mira_vs_reno_sdxl}
\end{figure}

Finally, to test scalability, we extend MIRA to SDXL~\citep{podell2023sdxl} and SDXL-Turbo~\citep{sauer2024adversarial} backbones using the first 50 HPDv2~\citep{hps} prompts, optimizing for HPSv2. In separate user studies, we compare our method with leading alternatives: DNO on SDXL and ReNO~\citep{eyring2024reno} on SDXL-Turbo. As shown in Figure~\ref{fig:user_study_winrate}, the results demonstrate a clear preference for our approach. MIRA achieves a 55.75\% win rate over DNO and a 63.35\% win rate over ReNO. This latter finding is supported by qualitative examples in Figure \ref{fig:mira_vs_reno_sdxl}), where MIRA tends to produce more literal and detailed compositions on the SDXL-Turbo backbone. Under equivalent sampling settings (same seed, default scheduler, single-step), MIRA produces cleaner, more literal interpretations with stable composition and style consistency (e.g., large-scale scene layout, totem-like geometry, and concept blending), whereas ReNO often introduces extra flourishes or alternative readings. These examples match our quantitative trends that MIRA delivers stronger alignment than noise-only regularization.

In summary, the evidence presented in this section provides a comprehensive answer to our second research question. Through large-scale automated evaluations, MIRA consistently outperforms a wide range of both fine-tuning and inference-time baselines. This strong quantitative performance is corroborated by focused human studies, where MIRA is preferred over the state-of-the-art DNO by a 4-to-1 margin on SDv1.5 and also wins decisively on the more powerful SDXL and SDXL-Turbo backbones. Together, these findings establish MIRA as a highly competitive method for inference-time alignment.

\subsection{How well can MIRA-DPO handle non-differentiable/black-box rewards?}
\label{experiments:q3}

Many real-world alignment objectives, such as human preferences or perceptual metrics, are inherently non-differentiable and cannot be directly optimized using gradient-based methods. To evaluate MIRA-DPO's effectiveness, we task it with a representative non-differentiable objective: maximizing JPEG compressibility. This is non-differentiable due to the quantization and rounding operations in the JPEG algorithm~\citep{Reich_2024, wallace1991jpeg}.

\begin{figure}[ht]
    \centering
    \includegraphics[width=\linewidth]{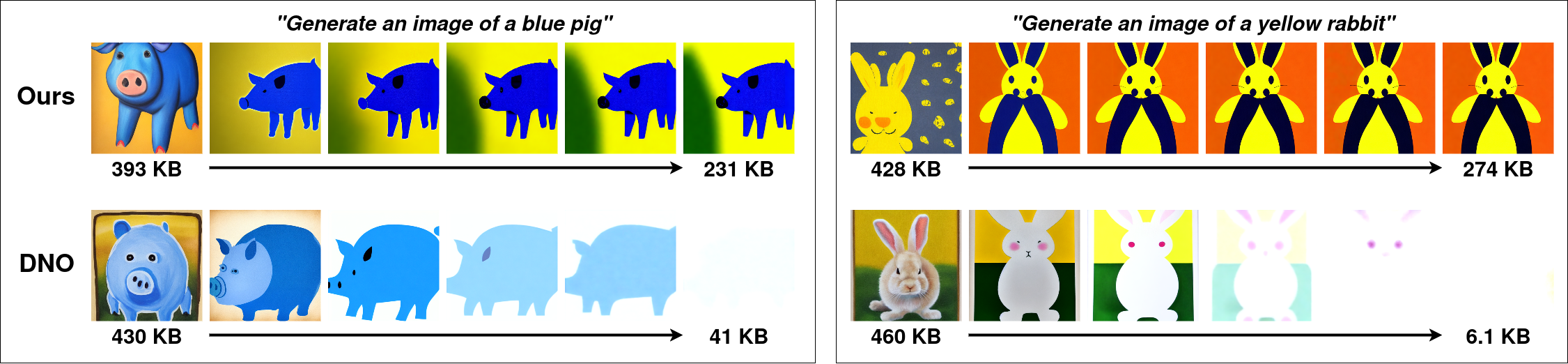}
    \vspace{-2mm}
    \caption{\textbf{MIRA-DPO preserves semantic content while optimizing a non-differentiable reward.} We compare MIRA-DPO with DNO on maximizing JPEG compressibility. We use prompts \textit{``generate an image of a blue pig''} (left) and \textit{``generate an image of a yellow rabbit''} (right), showing image evolution over 50 optimization steps. While DNO's output degrades into a nearly blank image, MIRA-DPO achieves significant compression while retaining the core features of the prompt.} 
    \label{fig:jpeg_compress}
\end{figure}

As shown in Figure \ref{fig:jpeg_compress}, while the baseline successfully minimizes file size, it also destroys the image's semantic content, noticeably hacking the reward. In contrast, MIRA-DPO's success in this task demonstrates that our distributional regularization is crucial for robustly handling black-box objectives, preventing noise optimization from taking shortcuts and ensuring the final image remains coherent. 

\subsection{Ablations and Additional Experiments}
\label{experiments:ablations}

In addition to win rate results shown in the above sections, we report the mean reward values for all methods (including MIRA) across all Animal prompts in Appendix \ref{appendix:additional_results} Table \ref{tab:rewards}. These values provide important context for interpreting trade-offs between reward maximization and prompt adherence for each objective (Aesthetic Score, HPSv2, PickScore, brightness, and darkness). Furthermore, we tune the hyperparameter $\beta$ to balance the trade-off between reward maximization and win rate. As shown in Appendix \ref{appendix:ablation_analysis} Figure \ref{fig:beta_ablation}, increasing $\beta$ systematically reduces reward scores but improves CLIPScore (quantifying image-prompt alignment). These ablations reveal that low $\beta$ tends to result in reward hacking, whereas high $\beta$ overconstrains, leading to an under-optimized reward.

Finally, for a holistic analysis, we compare MIRA to steering/sampling-based methods such as FK Steering~\citep{singhal2025generalframeworkinferencetimescaling} and CoDe~\citep{singh2025code}, matching MIRA's runtime with those of each baseline. All sampling methods run with their default configurations (SDv1.5 backbone) on one A100. For each method, we fix a single configuration, measure its mean per-prompt runtime $T$ on the Simple Animals prompts, and then run MIRA for the same budget $T$ on the same prompts. We report our head-to-head win rates for Aesthetic Score, darkness, and brightness rewards. This serves as a stress test where sampling-based methods can be sample-inefficient.

\begin{table}[ht]
    \centering
    \adjustbox{max width=\linewidth}{
    \begin{tabular}{l|c|c|c|c}
    \toprule
        Comparison & Aesthetic Score Win Rate (\%) $\uparrow$ & Darkness Win Rate (\%) $\uparrow$ & Brightness Win Rate (\%) $\uparrow$ & Peak GPU Mem (GB) $\downarrow$ \\
    \midrule
        MIRA vs FK Steering (16 Particles) & 57.78 & 75.56 & 73.33 & 20.55 \\
        MIRA vs DAS (8 Particles) & 37.78 & 77.78 & 80.00 & 23.90 \\
        MIRA vs CoDe ($N\!=\!20$) & 55.56 & 73.33 & 77.78 & 24.51 \\
        MIRA vs Best-of-$N$ ($N\!=\!50$) & 53.33 & 80.00 & 82.22 & 2.63 \\
    \bottomrule
    \end{tabular}}
    \vspace{2mm}
    \caption{\textbf{Comparison against state-of-the-art steering/sampling-based methods with matched wall-clock time budget.} For each baseline, the table lists peak GPU memory usage of each sampling-based method as well as MIRA's head-to-head win rates on Aesthetic Score, darkness, and brightness rewards. While MIRA's peak memory remains constant at 8.97GB, most steering-based competitors require over 20GB, highlighting a significant memory advantage for our method. Under a matched time budget, MIRA demonstrates highly competitive performance, particularly on brightness/darkness rewards where it consistently achieves win rates over 73\%.}
    \label{tab:same_compute_main}
\end{table}

\vspace{-5mm}
\section{Conclusions and Limitations}
\label{sec:discussion}

In this paper, we introduce MIRA, a distribution-regularized noise optimization method for inference-time alignment of diffusion models that mitigates reward hacking. By reformulating noise optimization as a constrained reward maximization problem, MIRA updates latent noise during sampling to improve reward while adhering to the user’s prompt. Extensive experiments demonstrate MIRA's strong empirical performance, with head-to-head win rates typically exceeding 60\% against baselines. Our work establishes that the key to mitigating such reward hacking is to directly regularize the output image distribution, which is more robust than prior noise-space constraints.

\noindent\textbf{Limitations.}  MIRA adds moderate inference-time compute beyond normal DDIM/DDPM sampling and relies on principled score-based surrogates, which may degrade on complex or out-of-distribution prompts; a formal analysis of the surrogate's theoretical tightness is an important direction for future work. While our score-based objective is sampler-portable, we focus our evaluation on DDIM for compute parity, as hyperparameters may vary by schedule. Improving MIRA's efficiency and combining it with sampling methods are left to future work.

%% file: sections/appendix.tex
\newpage
\appendix
\setcounter{table}{0}
\setcounter{figure}{0}

\tableofcontents
\clearpage
\section{Appendix}
\label{sec:appendix}

\subsection{Implementation Details}
\label{appendix:implementation_details}

We demonstrate results of MIRA on multiple reward functions, including Aesthetic Score, HPSv2, PickScore, and image brightness and darkness. The first three are human-aligned, whereas the last two are useful for easily visualizing reward hacking. We set $\beta=0.2$ for Aesthetic Score, $\beta = 0.5$ for HPSv2 and PickScore, $\beta = 1$ for brightness, and $\beta = 0.8$ for darkness. For JPEG compressibility, a non-differentiable reward, we use $\beta=1$. HPSv2 and PickScore use ViT-H/14 as the backbone, and Aesthetic Score uses ViT-L/14.

We implement MIRA with a learning rate of $0.01$ and the AdamW optimizer in all experiments. Our method runs on a single A100 Nvidia GPU. For DNO, we use noise regularization (i.e. PRNO) unless otherwise stated. Other baselines use their default configurations.

\textbf{User Study Details.} We obtain these results from anonymous volunteers who respond using an online form. We ask these participants: \textit{``Which of the below images would you personally prefer getting given the above prompt (based on your personal trade-off between prompt faithfulness and aesthetics)?''} We adopt this from ReNO. Given side-by-side images (prompt-matched, order-randomized), participants have the option to choose the left image, the right image, or neither/both. In the case of the last option, we assign a score of $0.5$ to that image pair. Due to computational restrictions, we limit the number of participants to 100. We further limit the number of images to 50 to respect the participants' time.

\subsection{Algorithms}
\label{appendix:algorithms}

For the reader's reference, we provide our loss for non-differentiable rewards as follows. Given prompt $c$,
\begin{equation}
    \label{eqn:ours_dpo_appendix}
    \mathcal{L}_{\text{MIRA-DPO}}=- \mathbb{E}_{x_0^{w},x_0^{l}} \log\!\sigma \Bigg(\!\beta\mathbb E_{\substack{x_{1:T}^{w}\sim p_\theta(\cdot|x_{0}^{w},z^w,c)\\x_{1:T}^{l}\sim p_\theta(\cdot|x_{0}^{l},z^l,c)}} \Bigg[\log \frac{p_{\theta}(x_{0:T}^{w}|z^w,c)}{p_{\theta}(x_{0:T}^{w}|z_0^w,c)} - \log \frac{p_{\theta}(x_{0:T}^{l}|z^l,c)}{p_{\theta}(x_{0:T}^{l}|z_0^l,c)}\Bigg]\!\Bigg).
\end{equation}

\begin{algorithm*}[ht]
\caption{MIRA for Non-differentiable Rewards}
\label{alg:mira_dpo}
\begin{algorithmic}[1]
\REQUIRE Initial noise $z_0^{(1)},z_0^{(2)}\sim\mathcal{N}(0, \mathbf{I})$, prompt $c$, reward function $r$, sampling process $G_\theta$, optimization steps $K_\text{opt}$, learning rate $\alpha$, regularization hyperparameter $\beta>0$

\STATE Initialize $z^{(1)} \gets z_0^{(1)}$, $z^{(2)} \gets z_0^{(2)}$
\FOR{$k=1, \dots, K_\text{opt}$} 
    \STATE $x_0^{(1)} \gets G_\theta(z^{(1)},c)$ \hfill \textit{// Generate images}
    \STATE $x_0^{(2)} \gets G_\theta(z^{(2)},c)$
    
    \STATE \textbf{if} $r(x_0^{(1)},c) > r(x_0^{(2)},c)$ \textbf{then} \hfill \textit{// Choose preferred image; assign winner and loser} \\ 
    \hspace{0.3cm} $x_0^w \gets x_0^{(1)}, x_0^l \gets x_0^{(2)}$ \\ 
    \hspace{0.3cm} $z^w\leftarrow z^{(1)}, z^l\leftarrow z^{(2)}$ \\ 
    \hspace{0.3cm} $z_0^w\leftarrow z_0^{(1)}, z_0^l\leftarrow z_0^{(2)}$ \\ 
    
    \STATE \textbf{else} \\ 
    \hspace{0.3cm} $x_0^w \gets x_0^{(2)}, x_0^l \gets x_0^{(1)}$ \\ 
    \hspace{0.3cm} $z^w\leftarrow z^{(2)}, z^l\leftarrow z^{(1)}$ \\
    \hspace{0.3cm} $z_0^w\leftarrow z_0^{(2)}, z_0^l\leftarrow z_0^{(1)}$

    \vspace{1mm}
    \STATE $\mathcal{L}_{\text{MIRA-DPO}} \gets \text{compute\_loss}(x_0^{w}, x_0^{l}, z^w, z^l, z_0^w, z_0^l)$ \hfill \textit{// Compute loss according to \cref{eqn:ours_dpo_appendix}}

    \STATE $z^{w} \gets z^{w} - \alpha\nabla_{\!z^{w}} \mathcal{L}_{\text{MIRA-DPO}}$ \hfill \textit{// Update noise vectors (also updating $z^{(1)}$ and $z^{(2)}$)} \\
    \STATE $z^{l} \gets z^{l} - \alpha\nabla_{\!z^{l}} \mathcal{L}_{\text{MIRA-DPO}}$
\ENDFOR
\RETURN $z^{w}$
\end{algorithmic}
\end{algorithm*}

\newpage
\subsection{Additional Qualitative Results}
\label{appendix:reward_hacking}

\textbf{Brightness reward hacking in MIRA vs. DNO}

Here, we show reward hacking on image brightness and darkness. We prompt the diffusion model with \textit{``generate an image of a black $[\textnormal{ANIMAL}]$''}. The contrast between the prompt and the reward provides a clear visualization of reward hacking.

\begin{figure}[ht]
     \centering
     \includegraphics[width=0.7\linewidth]{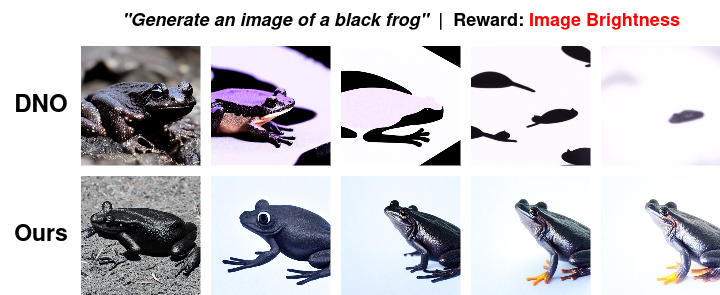}
     \caption{\textbf{MIRA vs. DNO in reward hacking}. On the image brightness reward, we demonstrate that MIRA is able to effectively mitigate reward hacking and generate better, more realistic images while maintaining prompt fidelity when compared to the state-of-the-art baseline. In the top row, after 50 optimization steps, DNO completely hacks the brightness reward and generates an image that is overly white and unrealistic. In contrast, in the bottom row, MIRA is able to mitigate reward hacking and produce much better images, while aligning with the target reward.}
     \label{fig:reward_hacking_appendix}
\end{figure}

\subsection{Additional Quantitative Results}
\label{appendix:additional_results}
In this section, we present additional results comparing MIRA to baselines.

\textbf{Win rates of all methods against SDv1.5} 

\begin{figure}[ht]
    \centering
    \includegraphics[width=\linewidth]{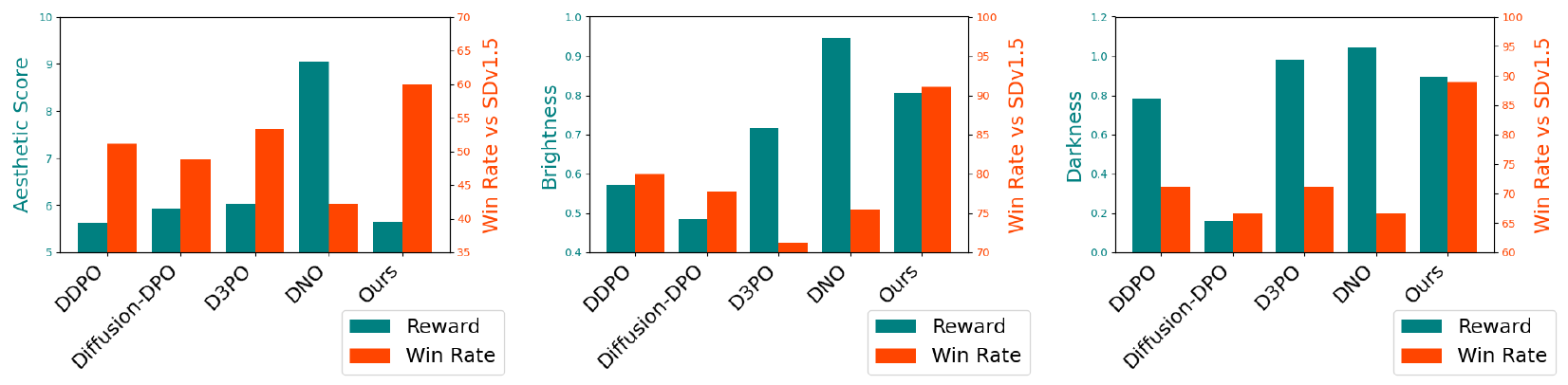}
    \vspace{-5mm}
    \caption{\textbf{Win rates (vs. SDv1.5) and average rewards across methods.} We evaluate our method, MIRA, on Aesthetic Score (left), brightness (middle), and darkness (right), comparing against DDPO, Diffusion-DPO, D3PO, and DNO. MIRA consistently outperforms baselines in win rates despite lower average rewards. Notably, higher rewards can indicate overoptimization, resulting in lower win rates.}
    \label{fig:reward_winrate_bar}
\end{figure}

In Table \ref{tab:winrate_vs_sd} below, we present the win rates of several methods, including MIRA, against SDv1.5 using Animal prompts. The results demonstrate that our method achieves the highest win rate in three rewards: Aesthetic Score, brightness, and darkness. We also represent this in the bar plot diagram in Figure~\ref{fig:reward_winrate_bar}. MIRA consistently outperforms baselines in win rates despite lower average rewards.

\begin{table}[h]
\centering
\begin{tabular}{c|ccc}
\toprule
Method (vs SDv1.5) & Aesthetic $\uparrow$ & Brightness $\uparrow$ & Darkness $\uparrow$ \\
\midrule
DDPO & 51.11 & 80.00 & 71.11\\
Diffusion-DPO & 48.89 & 77.77& 66.67\\
D3PO & 53.33 & 71.22 & 71.11\\
BoN ($N=30$) & 51.11 & 80.11 & 62.22\\
BoN ($N=40$) & 55.55 & 77.77 & 63.33\\
BoN ($N=50$) & 55.55 & 75.55 & 61.11\\
DNO & 42.22 & 75.56 & 66.67 \\
Ours & \textbf{60.00} & \textbf{91.11} & \textbf{88.89} \\
\bottomrule
\end{tabular}
\vspace{4mm}
\caption{\textbf{Win rates of all methods against SDv1.5.} The table reports the percentage of pairwise comparisons between all methods (including MIRA) vs SDv1.5 on three objectives: \textit{Aesthetic Score, Brightness,} and \textit{Darkness} with GPT-4o as the judge. We sample Best-of-$N$ (BoN) with base SDv1.5. Across all objectives, our method consistently achieves higher win rates than SDv1.5. These results demonstrate that our method not only outperforms previous baselines in terms of win rates over various target objectives, but is also able to generate better images while more effectively optimizing the target reward functions.}
\label{tab:winrate_vs_sd}
\end{table}

\newpage
\textbf{Overall rewards across methods and reward functions.} Table \ref{tab:rewards} presents the mean rewards of images generated by different methods (including MIRA). Rewards are averaged across all images generated from the Animal prompts.
\begin{table}[ht]
    \centering
     \adjustbox{max width=\linewidth}{
        \begin{tabular}{c|c|c|ccccc}
        \toprule
        Method & Training Time & Inference Time & Aesthetic & HPSv2 & PickScore & Brightness & Darkness \\
        \midrule
        SDv1.5 & - & 0.04 min & 5.367 & 0.278 & 20.390 & -0.123 & 0.087 \\
        DDPO & 12 h & - & 5.623 & 0.307 & 20.120 & 0.570 & 0.782\\
        Diffusion-DPO & 2 h & - & 5.930 & 0.302 & 21.086 & 0.485 & 0.159 \\
        D3PO & 24 h & - & 6.027 & 0.297 & 20.030 & 0.715 & 0.982\\
        BoN ($N = 30$) & - & 0.9 min & 6.044 & 0.318 & 21.404 & 0.006 & 0.329 \\
        BoN ($N = 40$) & - & 1.31 min & 6.080 & 0.320 & 21.427 & 0.094 & 0.339 \\
        BoN ($N = 50$) & - & 1.64 min & 6.111 & 0.321 & 21.365 & 0.105 & 0.353 \\
        InitNO & - & 
        0.5 min & 5.259 & 0.286 & 20.120 & -0.041 &
        0.113
        \\
        DyMO & - & 0.85 min &
        5.720 &
        0.322 &
        21.220 &
        0.320 &
        0.008 
        \\
        DNO & - & 5 min & 9.044 & 0.287 & 25.568 & 0.945 & 1.044 \\
        Ours & - & 5 min & 5.646 & 0.285 & 21.161 & 0.805 & 0.894 \\
        \bottomrule
        \end{tabular}
        }
    \vspace{2mm}
    \caption{\textbf{Overall rewards across methods and reward functions.} The above table compares our method with other baselines across various reward metrics and runtime. These rewards include Aesthetic Score, HPSv2, PickScore, brightness, and darkness. The leftmost column contains all methods. The second column (runtime) lists the approximate time needed to fine-tune or optimize each method (for a single image). We note DDPO, Diffusion-DPO, and D3PO take several hours due to fine-tuning. Though our method does not generate images with the highest rewards, we observe higher rewards compared to SDv1.5 after just five minutes of optimization with less reward hacking as discussed in the main paper. Inference times are measured using a single A100 GPU.}
    \label{tab:rewards}
\end{table}

\newpage
{\textbf{CLIPScores on the Simple Animals dataset.} We optimize several key methods (including fine-tuning methods) for Aesthetic Score on the Simple Animals dataset, evaluating CLIPScore as a prompt fidelity check. We observe that strong reward gains may come with weaker prompt adherence. Our method, MIRA, achieves the highest CLIPScore while remaining competitive on the actual target reward. We emphasize that \emph{these results should not be considered in a vacuum, and win rate is a more holistic metric.}}

\begin{table}[ht]
    \centering
    \begin{tabular}{c|cc}
    \toprule
        Method & CLIPScore & Aesthetic Score \\
    \midrule
        DDPO & 24.898 & 5.623 \\
        Diffusion-DPO & 25.340 & 5.930 \\
        D3PO & 23.298 & 6.027 \\
        BoN ($N=50$) & 24.070 & 6.044 \\
        InitNO & 24.555 & 5.720 \\
        DyMO & 25.397 & 5.720 \\
        DNO & 22.959 & 9.044 \\
        MIRA (Ours) & 25.964 & 5.646 \\
    \bottomrule
    \end{tabular}
    \vspace{4mm}
    \caption{\textbf{Comparison of CLIPScores according to method used, optimizing for Aesthetic Score.} We report CLIPScores, measuring prompt fidelity, and Aesthetic Score for various methods. MIRA achieves the best CLIPScore with comparable Aesthetic Score, whereas DNO achieves substantially higher Aesthetic Scores at the cost of prompt adherence.}
    \label{tab:clipscores}
\end{table}

\textbf{Hybrid feasibility (MIRA + FK Steering).} We implemented a simple hybrid combining MIRA with FK Steering to test complementarity. Concretely, we run multiple denoising trajectories in parallel and, at periodic intervals, select the candidate that maximizes the MIRA objective. The backbone remains frozen and configurations are fixed. This check is not wall-clock matched and serves only as a feasibility probe. On Simple Animals (45 prompts), optimizing for Aesthetic Score with $\beta=0.1$, \textbf{MIRA with FK Steering achieves a 55.56\% win rate over FK Steering alone (same backbone/GPU).}

\textbf{Trade-offs with compute.} We examine how reward changes as we vary the sampling budget while keeping the optimization procedure fixed. Specifically, we sweep the number of DDIM sampling steps (NFE) and compare MIRA to DNO under the same backbone and guidance. If the reward for a method increases significantly with additional sampling computation, this is indicative of reward hacking rather than genuine improvement. As shown in Fig. 4, DNO’s reward increases drastically with more DDIM steps, whereas MIRA remains comparatively stable. This is consistent with MIRA’s regularization, preventing the sampler from drifting toward high-score but unnatural solutions.

\begin{figure}[ht]
    \centering
    \includegraphics[width=0.6\linewidth]{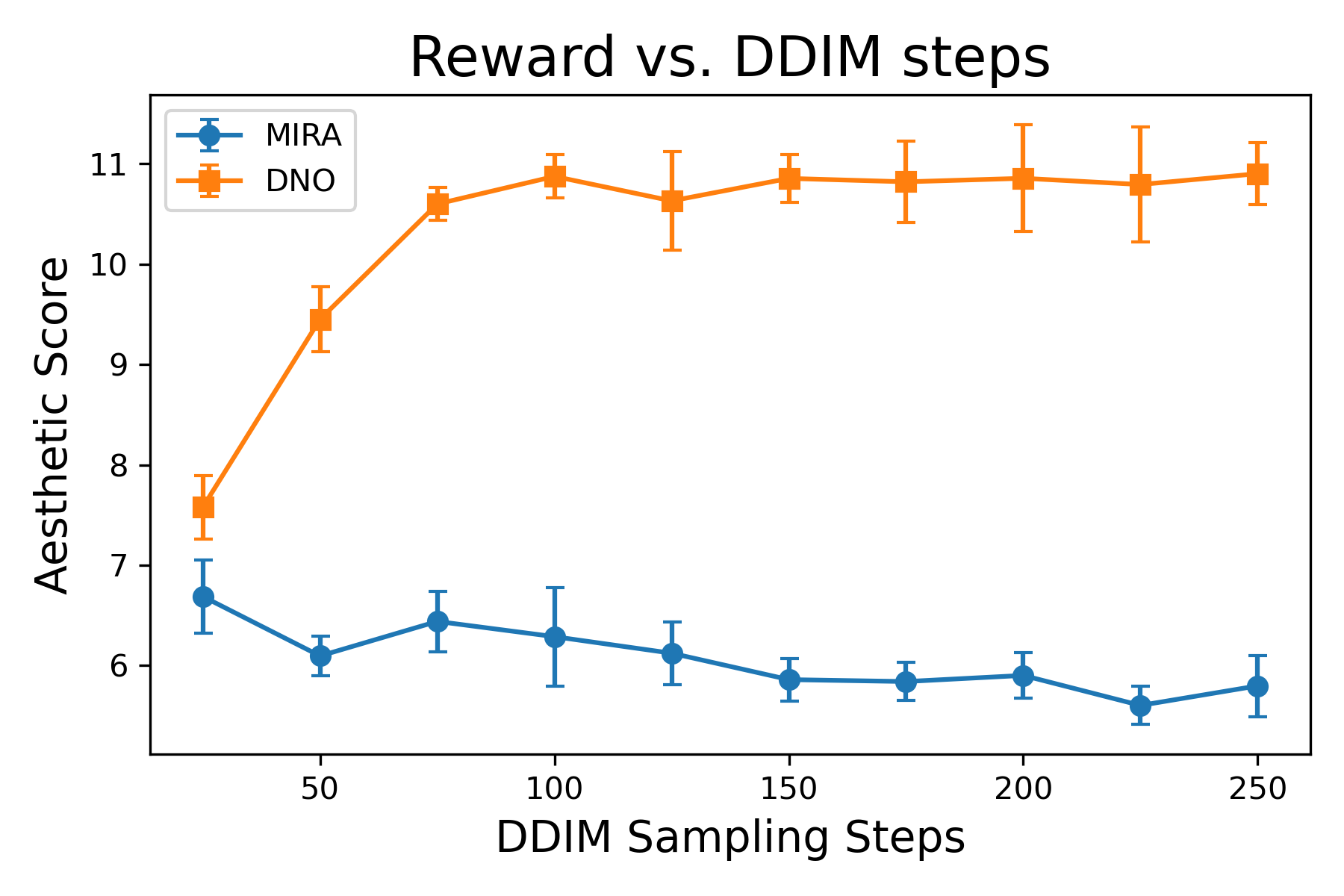}
    \vspace{-6mm}
    \caption{\textbf{Reward vs DDIM Steps (sampling compute) on Simple Animals dataset.} Mean Aesthetic Score vs. number of DDIM sampling steps ($\eta=1$, same backbone/CFG; optimization iterations fixed). DNO’s reward grows steadily with more steps, a hallmark of reward hacking; MIRA stays nearly flat (slight decline), indicating robustness to additional sampling compute. Error bars show variability across prompts. \textbf{The takeaway:} extra sampling compute disproportionately benefits noise-only optimization (DNO), while MIRA’s image-space regularization curbs this effect.}
    \label{fig:divergence_vs_compute}
\end{figure}

\newpage
\subsection{Ablation Analysis}
\label{appendix:ablation_analysis}

\textbf{Impact of $\beta$ on CLIPScores and Rewards} 

\begin{figure}[ht]
    \centering
    \includegraphics[width=1\linewidth]{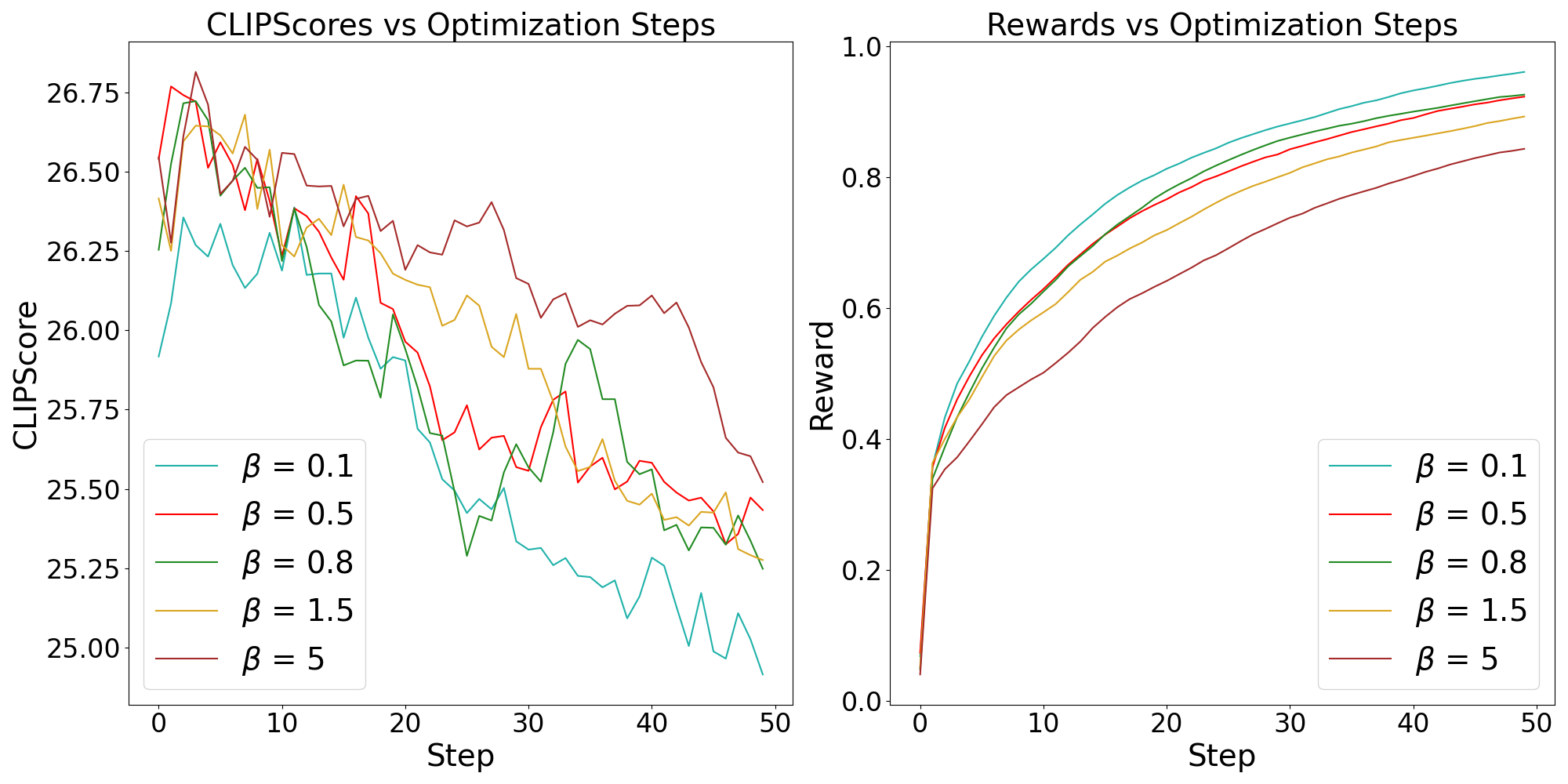}
    \vspace{-4mm}
    \caption{\textbf{Effect of different $\beta$ on CLIPScores (left) and rewards (right).} We obtain CLIPScores and rewards for all entries in the animals list in Appendix \ref{appendix:animal_prompts}. Curves depict the average result of five different seeds. The curves depict that lower $\beta$ exhibits higher rewards but lower CLIPScore. Higher $\beta$ tends to inadequately optimize for the reward.}
    \label{fig:beta_ablation}
\end{figure}

We justify our choice of hyperparameter $\beta$ by evaluating results across five different seeds, plotting the mean curves for CLIPScore and reward against optimization steps. CLIPScore is a metric that scores how well an image aligns with a given prompt. For our results, we use CLIPScore with a ViT-L/14 vision backbone and optimize for image darkness. As shown in Figure \ref{fig:beta_ablation}, increasing $\beta$ in our method leads to lower rewards but higher CLIPScores. We note that ``early stopping'' would yield higher CLIPScores but lower rewards.

%\newpage
\subsection{Animal Prompts List}
\label{appendix:animal_prompts}

We use the following set of Animal DDPO prompts.

\begin{center}
    \begin{tabular}{c|c|c|c|c|c|c|c|c}
        cat & dog & horse & monkey & rabbit & zebra & spider & bird & sheep \\
        deer & cow & goat & lion & tiger & bear & raccoon & fox & wolf \\
        lizard & beetle & ant & butterfly & fish & shark & whale & dolphin & squirrel \\
        mouse & rat & snake & turtle & frog & chicken & duck & goose & bee \\
        pig & turkey & fly & llama & camel & bat & gorilla & hedgehog & kangaroo
    \end{tabular}
\end{center}

\newpage
\subsection{GPT-4o Evaluation}
\label{appendix:gpt4_evaluation}

We compute the win rate of generated images from our method and baselines, employing GPT-4o as the judge. In all experiments, we use a temperature of 0 and Top-p of 1. The instruction prompt is as follows, where ``PROMPT'' and ``TARGET REWARD'' are placeholders.

\begin{tcolorbox}[title = System Prompt]
You are a helpful, impartial, and precise assistant tasked with evaluating two images based on a given text prompt, a reward criterion, and specific quality guidelines. The text prompt is: generate an image of a ``PROMPT''. The reward criterion is: ``TARGET REWARD.'' Please compare the two images and evaluate them based on the following:
\begin{enumerate}
    \item Adherence to the text prompt: Assess which image better satisfies the intent of the text prompt by being meaningful and directly relevant to the described subject. Do not use the reward criterion to determine adherence to the prompt.
    \item Satisfaction of the reward criterion: Determine which image more effectively meets the reward criterion.
\end{enumerate}

Choose the image that maximizes the reward without compromising adherence to the prompt.
With a lower priority, also ensure that: 
\begin{enumerate}
\itemsep0em 
    \item The image does not have overly vibrant colors or extreme sharpening.
    \item The lighting is balanced and natural. 
    \item The lighting is softer and more diffused. 
    \item The image appears realistic and natural, avoiding being overly saturated, sharpened, or monochromatic.
    \item Shapes and objects are not overly simplified, distorted, or stylized. 
    \item The image is not completely white or black, unless explicitly required. 
    \item Relevant textures are present, and the image is not overly simplistic or missing essential details. 
    \item The composition is dynamic, creating a sense of energy and movement.
\end{enumerate}

In the end, only output 1 if the first image is selected, or else output 2, and nothing else.

\textit{USER PROMPT}

\end{tcolorbox}

\subsection{Proof of Proposition 1}
\label{appendix:proof_prop}

\begin{proof}
To make our point that the change in the image distribution is unbounded with respect to the change in initial noise, we consider a simple setting with $z:=x_T$,  $\epsilon_T,\hdots,\epsilon_1\sim\mathcal N(0,\sigma^2\mathbf I)$, and the following random process
\begin{align}
    x_{t-1}&=\theta x_t+\epsilon_t, \nonumber
\end{align}
which holds from $t=T$ to $1$. Now, applying the recursive relation starting from $x_T=z$, we can write
\begin{align}
    x_{T-1}&=\theta z+\epsilon_T \nonumber \\
    x_{T-2}&=\theta x_{T-1}+\epsilon_{T-1}=\theta^2x_T+\theta\epsilon_T+\epsilon_{T-1} \nonumber \\
    &\vdots \nonumber \\
    x_0 &=\theta^Tz+\sum_{k=1}^T\theta^{k-1}\epsilon_{T-k+1}.
\end{align}
This implies that the mean of $x_0$ is
\begin{equation}\label{mean}
    \mathbb E\left[ x_0|z \right] = \theta^Tz,
\end{equation}
and variance of $x_0$ can be written as 
\begin{align}\label{variance}
    \text{Var}(x_0|z)&=\text{Var}\!\left( \sum_{k=1}^T\theta^{k-1}\epsilon_{T-k+1} \right) =\sum_{k=1}^T\theta^{2k-2}\sigma^2 \nonumber \\
    &=\sigma^2\frac{\theta^{2T}-1}{\theta^2-1},
\end{align}
 and for simplicity we assume $\left|\theta\right|>1$. From \cref{mean,variance}, we note that $x_0\sim \mathcal{N}\bigg(\theta^Tz, \sigma^2\frac{\theta^{2T}-1}{\theta^2-1}\bigg)$. Suppose that we have two noises $z_1,z_2\sim\mathcal N(0,\mathbf I)$. Hence, for two close noise $z_1$ and $z_2$, we can write 
\begin{equation}\label{final}
    d_\text{KL}\left[ p(\cdot|z_1)\|p(\cdot|z_2) \right]=\frac{\theta^{2T}{(\theta^2-1)}}{2\sigma^2(\theta^{2T}-1)} \cdot (z_1-z_2)^2,
\end{equation}
which holds from the closed-form expression of KL divergence between two Gaussian distributions. From the above expression in \cref{final}, we note that the rate of change is controlled by the term $\frac{\theta^{2T}{(\theta^2-1)}}{2\sigma^2(\theta^{2T}-1)}$, which is controlled by the norm of neural network parameter $\theta$ which can be arbitrarily large in practice. 

\end{proof}

\subsection{Derivation of the Practical Objective for Differentiable Rewards}
\label{appendix:proof_practical_algorithm}
Here, we derive the score function approximation of our Lagrangian objective,
\begin{align}
\label{eqn:ours_lagrangian_1}
\begin{split}
    \mathcal{J}_{\text{MIRA}}(z,c)=\mathbb{E}_{x_0 \sim p_{\theta}(\cdot|z,c)} \big[r(x_0,c)\big] - \beta d_{\text{KL}}\big[p_{\theta}(x_0|z,c)\ \| \ p_{\theta}(x_0|z_0,c)\big].
\end{split}
\end{align}

We consider the $d_{\text{KL}}$ term in \cref{eqn:ours_lagrangian_1}
\begin{align}
\label{eqn:eqn_1}
\begin{split}
    \mathcal{W} =  d_{\text{KL}}\big[p_{\theta}(x_0|z,c)\ \|\ p_{\theta}(x_0|z_0,c)\big].
\end{split}
\end{align}

We need to show that we can minimize $\mathcal W$ by minimizing
\begin{align}
\label{eqn:eqn_2}
\begin{split}
    \mathcal W'\approx\mathbb{E}_{\tau \sim p_{\theta}(\cdot|z,c)} \bigg[\sum_{t=0}^{T-1} \sigma_t^2\left(\|s(x_t|z_0,c)\|^2 - \|s(x_t|z,c)\|^2\right)\bigg],
\end{split}
\end{align}
with expectation over reverse trajectories $\tau$.

\vspace{5mm}
\begin{proof}
We start with~\cref{eqn:eqn_1} and write it as follows:
\begin{align}
\label{eqn:eqn_3}
\begin{split}
    \mathcal{W} = \mathbb{E}_{x_0 \sim p_{\theta}(\cdot|z,c)} \bigg[ \log \frac{p_{\theta}(x_0|z,c)}{p_{\theta}(x_0|z_0,c)}\bigg].
\end{split}
\end{align}

Note that $d_\text{KL}\big[p_\theta(x_0|z,c)\|p_\theta(x_0|z_0,c)\big]\leq d_\text{KL}\big[p_\theta(\tau|z,c)\|p_\theta(\tau|z_0,c)\big]$, establishing an upper bound. Therefore,

\begin{align}
\label{eqn:eqn_4}
     \mathcal{W} \leq \mathcal W'=\mathbb{E}_{\tau \sim p_{\theta}(\cdot|z,c)} \big[\log p_{\theta}(\tau|z,c) - \log p_\theta(\tau|z_0,c)\big].
\end{align}

By linearity of expectation, this is
\begin{align}
\label{eqn:eqn_4_expanded}
    \mathcal{W'} = \mathbb{E}_{\tau \sim p_{\theta}(\cdot|z,c)} \big[\log p_{\theta}(\tau|z,c)\big] - \mathbb{E}_{\tau \sim p_{\theta}(\cdot|z,c)}\big[\log p_{\theta}(\tau|z_0,c)\big].
\end{align}

Let us focus on the term $\mathbb{E}_{\tau \sim p_{\theta}(\cdot|z,c)} [\log p_{\theta}(\tau|z,c)]$, which we can expand as
\begin{align}
\label{eqn:eqn_5_product}
\begin{split}
    \mathbb{E}_{\tau \sim p_{\theta}(\cdot|z,c)} [\log p_{\theta}(\tau|z,c)] = \mathbb{E}_{\tau \sim p_{\theta}(\cdot|z,c)} \left[ \log \prod_{t=0}^{T-1} p_{\theta}(x_t|x_{t+1},z, c)\right].
\end{split}
\end{align}

We can expand \cref{eqn:eqn_5_product} as

\begin{align}
\label{eqn:eqn_5}
    &\mathbb{E}_{\tau \sim p_{\theta}(\cdot|z,c)} [\log p_{\theta}(\tau|z,c)] \nonumber
    \\&\hspace{1.5cm}= \mathbb{E}_{\tau \sim p_{\theta}(\cdot|z,c)} \Bigg[ \log \left (\exp \bigg(-\sum_{t=0}^{T-1} \frac{\|x_t - \mu_\theta(x_{t+1}, z,c)\|^2}{2\sigma_t^2}\bigg)\prod_{t=0}^{T-1}\frac{1}{\sigma_t\sqrt{2\pi}}\right)\Bigg]\nonumber
    \\&\hspace{1.5cm}= \mathbb{E}_{\tau \sim p_{\theta}(\cdot|z,c)} \Bigg[ -\sum_{t=0}^{T-1}\frac{\|x_t-\mu_\theta(x_{t+1},z,c)\|^2}{2\sigma_t^2}\Bigg]+\sum_{t=0}^{T-1}\log\frac{1}{\sigma_t\sqrt{2\pi}},
\end{align}

where $\mu_\theta(\cdot)$ is the predicted mean of the Gaussian at each step of the reverse process. We also know that for a Gaussian distribution,

\begin{align}
\label{eqn:eqn_6}
\begin{split}
     \|s(x_t|z,c)\| = \|\nabla_{x_t}\log p(x_t|z,c)\| \approx \|\nabla_{x_t} \log p_{\theta} (x_t|x_{t+1};z,c)\| ,
\end{split}
\end{align}
\begin{align}
\label{eqn:eqn_7}
\begin{split}
     \|s(x_t|z,c)\| \approx \left\|\nabla_{x_t} \log \left( \frac{1}{\sigma_t\sqrt{2\pi}} \exp \left( -\frac{\|x_t - \mu_\theta(x_{t+1}, t+1, z,c)\|^2}{2 \sigma_t^2} \right) \right) \right\|,
\end{split}
\end{align}
\begin{align}
\label{eqn:eqn_8}
\begin{split}
     \|s(x_t|z,c)\| \approx \left\|-\frac{x_t - \mu_\theta(x_{t+1}, t+1, z,c)}{\sigma_t^2}\right\|.
\end{split}
\end{align}

Replacing \cref{eqn:eqn_8} in \cref{eqn:eqn_5}, we get
\begin{align}
\label{eqn:eqn_9}
    \mathbb{E}_{\tau \sim p_{\theta}(\cdot|z,c)} [\log p_{\theta}(\tau|z,c)] \approx \mathbb{E}_{\tau \sim p_{\theta}(\cdot|z,c)} \Bigg[ -\sum_{t=0}^{T-1}\|s(x_t|z,c)\|^2\frac{\sigma_t^2}{2} \Bigg]+\sum_{t=0}^{T-1}\log\frac{1}{\sigma_t\sqrt{2\pi}}.
\end{align}

Applying \cref{eqn:eqn_9} to \cref{eqn:eqn_4}, we get 
\begin{align}
\label{eqn:eqn_4.1}
    \mathcal{W'} \approx \frac{1}{2}\mathbb{E}_{\tau \sim p_{\theta}(\cdot|z,c)}\left[ \sum_{t=0}^{T-1}\left(\|s(x_t|z_0,c)\|^2 - \|s(x_t|z,c)\|^2\right)\sigma_t^2 \right].
\end{align}

We can use \cref{eqn:eqn_4.1} in \cref{eqn:ours_lagrangian_1}. Our final objective is 
\begin{align}
\label{eqn:ours_practical_1}
\begin{split}
    \mathcal{J}_{\text{MIRA}}(z,c) \!=\! \mathbb{E}_{x_0 \sim p_{\theta}(\cdot|z,c)} \big[ r(x_0,c) \big] - \beta\mathbb{E}_{\tau \sim p_{\theta}(\cdot|z,c)} \bigg[\sum_{t=0}^{T-1} \sigma_t^2\left(\|s(x_t|z_0,c)\|^2 - \|s(x_t|z,c)\|^2\right)\bigg].
\end{split}
\end{align}
\end{proof}

\subsection{Derivation of DPO Objective}
\label{appendix:dpo_derivation}

We begin with our original objective,
\begin{equation}
    \max_{z}\mathcal{J}_{\text{MIRA}}(z,c) = \max_{z}\left[\mathbb{E}_{x_0 \sim p_{\theta}(\cdot|z,c)} \big[r(x_0,c)\big]- \beta d_{\text{KL}}\left[p_{\theta}(x_0|z,c)\ \! \| \! \ p_{\theta}(x_0|z_0,c)\right]\right].
\end{equation}

Let $R(x_{0:T},c)$ be the reward along the whole sampling trajectory such that 
\begin{equation}
    r(x_0,c):=\mathbb E_{x_{1:T}\sim p_\theta(\cdot|x_0,z,c)}\left[R(x_{0:T},c)\right].
\label{eqn:r_to_R}
\end{equation}

This means
\begin{align}
    \max_{z}\mathcal{J}_{\text{MIRA}}(z,c)&=\max_z\left[ \mathbb E_{\substack{x_0\sim p_\theta(\cdot|z,c)\\x_{1:T}\sim p_\theta(\cdot|x_0,z,c)}}\left[ R(x_{0:T},c) \right] -\beta d_\text{KL}\left[p_{\theta}(x_0|z,c)\ \! \| \! \ p_{\theta}(x_0|z_0,c)\right] \right]\nonumber \\
    &=\min_{z}\left[-\mathbb{E}_{\substack{x_0 \sim p_{\theta}(\cdot|z,c)\\x_{1:T}\sim p_\theta(\cdot|x_0,z,c)}} \left[R(x_{0:T},c)\right]+ \beta d_{\text{KL}}\left[p_{\theta}(x_0|z,c)\ \! \| \! \ p_{\theta}(x_0|z_0,c)\right]\right]\nonumber \\
    &=\min_{z}\left[-\mathbb{E}_{x_0 \sim p_{\theta}(\cdot|z,c)} \bigg[\mathbb E_{x_{1:T}\sim p_\theta(\cdot|z,c)}\left[R(x_{0:T},c)\right]-\beta \log\frac{p_\theta(\cdot|z,c)}{p_\theta(\cdot|z_0,c)}\bigg]\right]\nonumber \\
    &=\min_{z}\left[\mathbb{E}_{x_0 \sim p_{\theta}(\cdot|z,c)} \bigg[\log\frac{p_\theta(\cdot|z,c)}{p_\theta(\cdot|z_0,c)} - \frac{1}{\beta}\mathbb E_{x_{1:T}\sim p_\theta(\cdot|x_0,z,c)}\left[R(x_{0:T},c)\right] \bigg]\right]\nonumber \\
    &\leq\min_{z}\left[\mathbb{E}_{x_{0:T} \sim p_{\theta}(\cdot|z,c)} \bigg[\log\frac{p_\theta(\cdot|z,c)}{p_\theta(\cdot|z_0,c)} - \frac{1}{\beta}R(x_{0:T},c) \bigg]\right].
\end{align}

\begin{align}
    \max_{z}\mathcal{J}_{\text{MIRA}}(z,c) &\leq \min_{z} \left[ \mathbb E_{x_{0:T}\sim p_\theta(\cdot|z,c)}\log\frac{p_\theta(\cdot|z,c)}{\frac{1}{Z(c)}p_\theta(\cdot|z_0,c)\exp(\frac{1}{\beta}R(x_{0:T},c))}-\log Z(c) \right], \\
    & \hspace{0.4cm} \text{where \hspace{0.2cm} } Z(c) = \sum_{x_0} p_\theta(\cdot|z_0,c)\exp(\frac{1}{\beta}R(x_{0:T},c)). \\
    \max_{z}\mathcal{J}_{\text{MIRA}}(z,c) &\leq\min_{z}\left[ \mathbb E_{x_{0:T}\sim p_\theta(\cdot|z,c)}\log\frac{p_\theta(\cdot|z,c)}{p_\theta(\cdot|z^*,c)}-\log Z(c) \right],\\
    & \hspace{0.4cm} \text{where \hspace{0.2cm} } p_\theta(\cdot|z^*,c) = \frac{1}{Z(c)}p_\theta(\cdot|z_0,c)\exp(\frac{1}{\beta}R(x_{0:T},c)). \\
    \max_{\mathbf z}\mathcal{J}_{\text{MIRA}}(z,c) &\leq\min_{z}\left[ d_{\text{KL}}[p_\theta(x_{0:T}|z,c)\ \|\ p_\theta(x_{0:T}| z^*,c)]-\log Z(c) \right].
\end{align}
Since $Z(c)$ does not depend on $z$, the minimum is achieved when the KL term is zero, which holds when the two distributions are identical, i.e.,
\begin{align}
\label{eqn:eqn_12}
    p_\theta(x_{0:T}|z^*,c) = \frac{1}{Z(c)}p_\theta(x_{0:T}|z_0,c)\exp(\frac{1}{\beta}R(x_{0:T},c)).
\end{align}
Taking the logarithm on both sides and rearranging the terms, we get
\begin{align}
\label{eqn:eqn_reward}
    R(x_{0:T},c) = \beta \log Z(c) + \beta \log \frac{p_{\theta}(x_{0:T}|z,c)}{p_{\theta}(x_{0:T}|z_0,c)}.
\end{align}
This means, from \cref{eqn:r_to_R},
\begin{equation}
    r(x_0,c)=\beta\log Z(c)+\beta\mathbb E_{x_{1:T}\sim p_\theta(\cdot|x_0,z,c)}\left[\log\frac{p_\theta(x_{0:T}|z,c)}{p_\theta(x_{0:T}|z_0,c)}\right].
\label{eqn:eqn_reward_2}
\end{equation}

Assume that we have access to a dataset $\{(z^{(1)},x_0^{(1)}),(z^{(2)},x_0^{(2)}),c\}_{i=1}^{N}$, where the generated image $x_0^{(1)}$ is preferred over $x_0^{(2)}$. Under the Bradley-Terry model, the human preference distribution can be written as:
\begin{align}
\label{eqn:bt_model_1}
    P^*(x_0^{(1)}\succ x_0^{(2)}|z^{(1)},z^{(2)},c) & =\frac{\exp(r^*(x_0^{(1)},c))}{\exp(r^*(x_0^{(1)},c))+\exp(r^*(x_0^{(2)},c))} \nonumber
    \\ & = \sigma (r^*(x_0^{(1)},c) - r^*(x_0^{(2)},c)).
\end{align}

Framing the problem as a binary classification, we have the negative log-likelihood loss as:
\begin{align}
\label{eqn:dpo_1}
    \mathcal{L}_{\text{MIRA-DPO}}(z^{(1)},z^{(2)},c) = - \mathbb{E}_{\substack{x_0^{(1)} \sim p_{\theta}(\cdot|z^{(1)},c) \\ x_{0}^{(2)} \sim p_{\theta}(\cdot|z^{(2)},c)}} [\log \sigma (r^*(x_0^{(1)},c) - r^*(x_0^{(2)},c))].
\end{align}

Using~\cref{eqn:eqn_reward_2} in~\cref{eqn:dpo_1}, we get
\begin{equation}
\begin{split}
    &\mathcal{L}_{\text{MIRA-DPO}}(z^{(1)},z^{(2)},c) = \\&-\!\mathbb{E}_{\substack{x_0^{(1)} \sim p_{\theta}(\cdot|z^{(1)},c) \\ x_{0}^{(2)} \sim p_{\theta}(\cdot|z^{(2)},c)}} \log \sigma \Bigg(\beta\mathbb E_{\substack{x_{1:T}^{(1)}\sim p_\theta(\cdot|x_0^{(1)},z^{(1)},c)\\x_{1:T}^{(2)}\sim p_\theta(\cdot|x_0^{(2)},z^{(2)},c)}} \left[\log \frac{p_{\theta}(x_{0:T}^{(1)}|z^{(1)},c)}{p_{\theta}(x_{0:T}^{(1)}|z_0^{(1)},c)} - \log\frac{p_{\theta}(x_{0:T}^{(2)}|z^{(2)},c)}{p_{\theta}(x_{0:T}^{(2)}|z_0^{(2)},c)}\right]\Bigg).
\label{eqn:dpo_obj}
\end{split}
\end{equation}

Since we assume $x_0^{(1)}\succ x_0^{(2)}$, we know $x_0^w:=x_0^{(1)}$ and $x_0^l:=x_0^{(2)}$. We can rewrite \cref{eqn:dpo_obj} as
\begin{equation}
\begin{split}
    &\mathcal{L}_{\text{MIRA-DPO}}(z^{w},z^{l},c) = \\&-\!\mathbb{E}_{\substack{x_0^{w} \sim p_{\theta}(\cdot|z^{w},c) \\ x_{0}^{l} \sim p_{\theta}(\cdot|z^{l},c)}} \log \sigma \Bigg(\beta\mathbb E_{\substack{x_{1:T}^{w}\sim p_\theta(\cdot|x_0^{w},z^{w},c)\\x_{1:T}^{l}\sim p_\theta(\cdot|x_0^{l},z^{l},c)}} \left[\log \frac{p_{\theta}(x_{0:T}^{w}|z^{w},c)}{p_{\theta}(x_{0:T}^{w}|z_0^{w},c)} - \log\frac{p_{\theta}(x_{0:T}^{l}|z^{l},c)}{p_{\theta}(x_{0:T}^{l}|z_0^{l},c)}\right]\Bigg),
\end{split}
\end{equation}

which can be conceptually understood as the standard DPO loss, $\mathcal L=-\mathbb E[\log\sigma(\hat r^w - \hat r^l)]$. Here, we define the implicit, regularized MIRA reward $\hat r$ as the log-ratio of the noise-optimized and reference model probabilities, which is practically realized using our score-based surrogate. Hence, this is our formulation in the main paper.

% ---
\vspace{5mm}
\textbf{Practically, we can use our score function surrogate.}

By Jensen's inequality,
\begin{equation}
    \mathcal{L}_{\text{MIRA-DPO}}(z^{w},z^{l},c) \leq -\!\log \sigma \Bigg(\beta\mathbb E_{\substack{x_{0:T}^{w}\sim p_\theta(\cdot|z^{w},c)\\x_{0:T}^{l}\sim p_\theta(\cdot|z^{l},c)}} \Bigg[\log \frac{p_{\theta}(x_{0:T}^{w}|z^{w},c)}{p_{\theta}(x_{0:T}^{w}|z_0^{w},c)} - \log\frac{p_{\theta}(x_{0:T}^{l}|z^{l},c)}{p_{\theta}(x_{0:T}^{l}|z_0^{l},c)}\Bigg]\Bigg).
\end{equation}
Therefore,
\begin{equation}
\begin{split}
    \mathcal{L}_{\text{MIRA-DPO}}(z^{w},z^{l},c) &\leq -\!\log\sigma\bigg(\beta d_\text{KL}\left[p_\theta(x_{0:T}^{w}|z^{w},c)\ \|\ p_\theta(x_{0:T}^{w}|z_0^{w},c)\right]\nonumber\\&\hspace{3.5cm}-\beta d_\text{KL}\left[p_\theta(x_{0:T}^{l}|z^{l},c)\ \|\ p_\theta(x_{0:T}^{l}|z_0^{l},c)\right] \bigg).
\end{split}
\end{equation}

As we have shown previously in \ref{appendix:proof_practical_algorithm}, we can approximate the KL terms. Our final loss is
\begin{equation}
\begin{split}
    &\mathcal{L}_{\text{MIRA-DPO}}(z^{w},z^{l},c)\\&\hspace{2cm}=-\!\log\sigma\bigg(\beta\mathbb{E}_{x_{0:T}^{w} \sim p_{\theta}(\cdot|z^{w},c)} \bigg[\sum_{t=0}^{T-1} \sigma_t^2\left(\|s(x_t^{w}|z_0^{w},c)\|^2 - \|s(x_t^{w}|z^{w},c)\|^2\right)\bigg]\\
    &\hspace{3cm}-\beta\mathbb{E}_{x_{0:T}^{l} \sim p_{\theta}(\cdot|z^{l},c)} \bigg[\sum_{t=0}^{T-1} \sigma_t^2\left(\|s(x_t^{l}|z_0^{l},c)\|^2 - \|s(x_t^{l}|z^{l},c)\|^2\right)\bigg]\bigg).   
\end{split}
\label{eqn:dpo_final}
\end{equation}